\documentclass[10pt]{article}

\usepackage{amsfonts}
\usepackage{amsmath}
\usepackage{amssymb}

\usepackage{overpic}
\usepackage{bm}
\usepackage{multirow}

\usepackage{amsthm}
\usepackage{sidecap}

\usepackage{algorithm2e}

\usepackage{stfloats}

\newcommand{\mat}[1]{\ensuremath{\mathbf{#1}}}

\def\dictm{\mat{D}}

\def\datam{\mat{X}}
\def\datav{\mat{x}}

\def\coefm{\mat{Z}}
\def\coefv{\mat{z}}

\def\weight{\lambda}

\def\ndims{m}
\def\natoms{q}

\def\nsamples{n}

\def\groupset{\mathcal{G}}
\def\ngroups{|\mathcal{G}|}

\def\F{\mathcal{F}}
\def\L{\mathcal{L}}
\def\T{\mathcal{T}}

\def\gi{r} 
\def\ai{i} 

\def\reals{\ensuremath{\mathbb{R}}}
\def\naturals{\ensuremath{\mathbb{N}}}

\newcommand{\norm}[1]{\ensuremath{\left\|#1\right\|}}    
\newcommand{\setdef}[1]{\ensuremath{\{#1\}}}

\def\sgn{\mathrm{sign}} 

\newcommand{\prox}{\operatorname{\bb{\pi}}}

\newcommand{\argmin}[1]{\underset{#1}{\operatorname{argmin}}}

\newcommand{\bb}[1]{\bm{\mathrm{#1}}}
\newcommand{\bi}[1]{\bm{{#1}}}

\def\Tr{\mathrm{T}}

\newcommand{\citep}[1]{\cite{#1}}

\def\enc{\bi{z}}
\def\dec{\bi{x}}

\newtheorem{theorem}{Theorem}
\newtheorem{corollary}{Corollary}

\begin{document}

\title{Learning Efficient Sparse and Low Rank Models}

\author{P. Sprechmann,\thanks{P. Sprechmann and G. Sapiro are with the Department of Electrical and Computer Engineering, Duke University, Durham 27708, USA. Email: pablo.sprechmann@duke.edu, guillermo.sapiro@duke.edu.} A. M. Bronstein,\thanks{A. M. Bronsteind is with School of Electrical Engineering, Tel Aviv University, Tel Aviv 69978, Israel.Email: bron@eng.tau.ac.il.} and G. Sapiro$^\ast$\thanks{Work partially supported by NSF, ONR, NGA, DARPA, AFOSR, ARO, and BSF.}
}

\maketitle

\begin{abstract}
Parsimony, including sparsity and low rank, has been shown to successfully model data in numerous machine learning and signal processing tasks. Traditionally, such modeling approaches rely on an iterative algorithm that minimizes an objective function with parsimony-promoting terms. The inherently sequential structure and data-dependent complexity and latency of iterative optimization constitute a major limitation in many applications requiring real-time performance or involving large-scale data. Another limitation encountered by these modeling techniques is the difficulty of their inclusion in discriminative learning scenarios. In this work, we propose to move the emphasis from the model to the pursuit algorithm, and develop a \emph{process-centric} view of parsimonious modeling, in which a learned deterministic fixed-complexity pursuit process is used in lieu of iterative optimization.  We show a principled way to construct learnable pursuit process architectures for structured sparse and robust low rank models, derived from the iteration of proximal descent algorithms. These architectures learn to approximate the exact parsimonious representation at a fraction of the complexity of the standard optimization methods. We also show that appropriate training regimes allow to naturally extend parsimonious models to discriminative settings. State-of-the-art results are demonstrated on several challenging problems in image and audio processing with several orders of magnitude speedup compared to the exact optimization algorithms.

\end{abstract}

\section{Introduction}
\label{sec.introduction}

Parsimony, preferring a simple explanation to a more complex one, is probably one of the most intuitive principles widely adopted in the modeling of nature. The past two decades of research have shown the power of parsimonious representation in a vast variety of applications from diverse domains of science.

One of the simplest among parsimonious models is \emph{sparsity}, asserting that the signal has many coefficients close or equal to zero when represented in some domain, usually referred to as \emph{dictionary}. The pursuit of sparse representations was shown to be possible using tools from convex optimization, in particular,
via $\ell_1$ norm minimization \citep{CDS99,tibshirani96}.
%
Works \cite{olshausen1996emergence,aharon2006img}, followed by many others, introduced efficient computational techniques for dictionary learning and adaptation. Sparse modeling is in the heart of modern approaches to image enhancement such as denoising, demosaicing, impainting, and super-resolution, to mention just a few.

As many classes of data are not described well by the element-wise sparsity model and the $\ell_1$ norm inducing it, more elaborate structured sparse models have been developed, in which non-zero elements are no more unrelated, but appear in groups or hierarchies of groups \citep{yuan06,jacob2009group,zhao-2009-37,JenattonMOB11,JournalHiLasso,peleg2012exploiting}. Such models have been shown useful in the analysis of functional MRI and genetic data  for example.

In the case of matrix-valued data, complexity is naturally measured by the rank, which also induces a notion of parsimony. A recent series of works have elucidated the beautiful relationship between sparsity and low rank representations, showing that rank minimization can be achieved through convex optimization \cite{srebro2005rank,candes2009exact}. The combination of low-rank and sparse models paved the path to new robust alternatives of principal component analysis (RPCA) \citep{Candes2011-JACM,xu2012robust} and nonnegative matrix factorization (RNMF) \citep{zhang2011robust}, and addressing challenging matrix completion problems \citep{candes2009exact}. RPCA was also found useful in important applications such as face recognition and modeling, background modeling, and audio source separation. 
Another relevant low rank modeling scheme is non-negative matrix factorization (NMF)\cite{NMF},
where the input vectors are represented as non-negative linear combination of a non-negative under-complete
dictionary. NMF has been particularly successful in applications such as object recognition and audio processing. %

\subsection{From model-centric to process-centric parsimonious modeling}

All existing parsimonious modeling methods essentially follow the same pattern: First, an objective comprising a data fitting term and parsimony-promoting penalty terms is constructed; next, an iterative optimization algorithm is invoked to minimize the objective, pursuing either the parsimonious representation of the data in a given dictionary, or the dictionary itself. Despite its remarkable achievements, such a \emph{model-centric} approach suffers from critical disadvantages and limitations. 

The inherently sequential structure and the data-dependent complexity and latency of iterative optimization tools often constitute a major computational bottleneck. The quest for efficiently solving sparse representation pursuit has given rise to a rich family of algorithms, both for sparse coding \cite{daubechies2004iterative,fista,Osher,nesterov07,bertsekas1999nonlinear,Bach11}
, RPCA \citep{Candes2011-JACM,CaiCS10,MuDYY11,Recht:2011wv} and NMF \citep{NMF,Lin07projectedgradient} problems.
%
%
%
%
Despite the permanent progress reported in the literature, the state-of-the-art algorithms require hundreds or thousands of iterations to converge, making their use impractical in scenarios demanding real-time performance or involving large-scale data.


Relying on the explicit solution of an optimization problem furthermore limits the applicability of parsimonious models in supervised learning scenarios, where the higher-level training objective would depend on the solution of the lower-level pursuit problem. The resulting bilevel optimization problems are notoriously difficult to solve in general; the non-differentiability of the lower-level parsimony-inducing objective makes the solution practically impossible \citep{colson2007overview}. This partially explains why sparse representations, that are so widely adopted for the construction of \emph{generative} models, had such a modest success in the construction of \emph{discriminative} models.

In this paper, we take several steps to depart from the model-centric ideology relying on an iterative solver
by shifting the emphasis from the model to the \emph{pursuit process}. 
%
Our approach departs from the observation that, despite being highly non-linear and hard to compute, the mapping between a data vector and its parsimonious representation resulting from the optimization procedure is deterministic. The curse of dimensionality precludes the approximation of this mapping on all possible, even modestly sized input vectors; however, since real data tend to have low intrinsic dimensionality, the mapping can be inferred explicitly on the support of the distribution of the input data.

Recently, \cite{jarrett2009best,kavukcuoglu2010fast} have proposed to trade off precision in the sparse representation for computational speed-up
by learning non-linear regressors capable of producing good approximations of sparse codes in a fixed amount of time.
However, the large number of degrees of freedom, for which a good initialization is difficult to provide, made this effort only modestly successful. In their inspiring recent paper, \cite{LecunNN} showed that a particular network architecture can be derived from the iterative shrinkage-thresholding (ISTA) \citep{daubechies2004iterative} and proximal coordinate descent (CoD) algorithms \citep{Osher}.

These works were among the first to bridge between the optimization based sparse models
and the inherently process-centric neural networks, and in particular auto-encoder networks \cite{hinton2006,autoencoders}, extensively explored
by the deep learning community.

\subsection{Contributions}

The main contribution of this paper is to propose a comprehensive framework for process-centric parsimonious modeling.
The obtained encoders can be used to produce fast approximants or predictors of optimization based parsimonious models or as modelers in their own right,
this is, pursuit processes that might not be minimizing any specific objective function.
Specifically, this paper makes four main contributions:

First, in Section~\ref{sec:learning}, we propose a process-centric approach to parsimonious modeling. We begin by proposing a principled way to construct encoders capable of approximating an important family of parsimonious models (briefly described in Section~\ref{sec.sparse.models}), including general sparse coding paradigms (hierarchical and non-overlapping group sparsity), robust PCA, and NMF. By extending the original ideas in \cite{LecunNN}, we propose tailored pursuit architectures derived from first-order proximal descent algorithms, which are briefly presented in Section~\ref{sec:prox}. Note that unlike the standard sparse coding setting, the exact first-order RPCA and RNMF algorithms cannot be used directly, as each iteration involves the singular value decomposition (SVD). As a remedy, we propose to use an algorithm inspired by the non-convex optimization techniques in \cite{Recht:2011wv}.

Second, this new approach allows the encoders to be trained in an online manner, which makes the fast encoders no more restricted to work with a fixed distribution of input vectors known \emph{a priori} (limitation existing, for example, in \cite{LecunNN}), and removes the need to run the exact algorithms at training.
The proposed approach can be used with a predefined dictionary or learn it in an online manner on the very same data vectors fed to it.
While differently motivated, in this setting, our framework is related to the sparse autoencoders \cite{autoencoders}. 

Third, we show that abandoning the iterative minimization in favor of a learned pursuit process allows to incorporate
the parsimonious representation into higher-level optimization problems in a natural way. In particular,  in Section~\ref{sec:training.regimes} we show a very simple and efficient
extension of the proposed RPCA framework to cases where the data undergo an unknown transformation that is sought for
during the pursuit \citep{Peng2011-PAMI}. We also show the construction of discriminative parsimonious models.

Finally, in Secion~\ref{sec.experiments} we demonstrate our approaches on applications in image classification, face modeling, signal separation and denoising, and speaker identification, where our fast encoders perform similarly to or better than the iterative pursuit processes at a fraction of the complexity of the latter. Faster than real-time state-of-the-art results are achieved in several such applications.
The present paper generalizes and gives a more rigorous treatment to results previously published by the authors in \cite{ICML,SBS_ismir}.


\section{Parsimonious models}
\label{sec.sparse.models}

Let $\datam \in \reals^{\ndims \times \nsamples}$ be a give data matrix. In this work, we concentrate our attention on the general \emph{parsimonious modeling} problem that can be posed as the solution of the minimization problem
\begin{eqnarray}
\min_{}
\frac{1}{2}\norm{\datam-\dictm\coefm}_\mathrm{F}^2 +\psi(\coefm) + \phi(\dictm), \label{eq:parsimonious.modeling}
\end{eqnarray}
optimized over $\coefm\in\reals^{\natoms \times \nsamples}$ alone or jointly with $\dictm \in \reals^{\ndims\times \natoms}$. Here $\coefm \in \reals^{\natoms  \times \nsamples}$ is the representation (parsimonious code) of the data in the dictionary, and the penalty terms
$\psi(\coefm)$ and $\phi(\dictm)$ induce a certain structure of the code and the dictionary, respectively. When the minimization is performed over both the dictionary and the representation, it is non-convex.

We will explicitly distinguish between parsimonious \emph{coding} or \emph{representation pursuit} problems (representing data with a given model), and the harder parsimonious \emph{modeling} problem (constructing a model describing given data, e.g., learning a dictionary).
In the former, $\dictm$ is fixed  and $\phi(\dictm)$ is constant. Most useful formulations use convex regularization $\psi(\coefm)$ of the representation.

In many relevant applications, the entire data matrix $\datam$ is not available \emph{a priori}.
The data samples $\{\bb{x}_t\}_{t\in \naturals}$, $\bb{x}_t\in \reals^{m}$,  arrive sequentially;
the index $t$ should be interpreted as time. \emph{Online parsimonious modeling}
aims at estimating and refining the model as the data come in \cite{mairal2009online}.
The need for online schemes also arises when the available training data are simply too large to be handled together.
When the regularized function $\psi$ is vector-wise separable,
\begin{eqnarray}
\psi(\coefm) &=& \sum_{i=1}^{\nsamples} \psi(\coefv_i),
\end{eqnarray}
problem \eqref{eq:parsimonious.modeling} can solved in an online fashion using an alternating minimization scheme.
As a new data vector $\bb{x}_t$ is received, we first obtain its representation $\bb{z}_t$ given the current model estimate, $\bb{D}_{t-1}$.
This is achieved by solving the representation pursuit problem
\begin{eqnarray}
\bb{z}_t &=&\argmin{\bb{z}}
\frac{1}{2} \norm{\datav_t - \dictm_{t-1}\coefv}^2_2 + \psi(\coefv).
\label{eq:parsimonious.modeling.online}
\end{eqnarray}
Then, we update the model using the coefficients, $\{\bb{z}_j\}_{j\leq t}$, computed during the previous steps of the algorithm,
\begin{eqnarray}
\dictm_t &=& \argmin{\dictm} \sum\limits_{j=1}^{t} \beta_j \frac{1}{2} \norm{\bb{x}_j - \dictm \coefv_j}^2_2 +\phi(\dictm),
\label{ec.convex.unconstrained.US.online.dict}
\end{eqnarray}
where  $\beta_j\in [0,1]$ is a forgetting factor that can be added to rescale older information so that newer estimates have more weight. This can be efficiently solved without remembering all the past codes \cite{mairal2009online}.

In what follows, we detail several important instances of \eqref{eq:parsimonious.modeling}, for both modeling or pursuit, and how they can be
cast as online learning problems.

\subsection{Structured sparsity}

The underlying assumption of sparse models is that the input vectors can be reconstructed accurately as a linear combination of the dictionary
atoms with a small number of non-zero coefficients.
Sparse models are enforced by using sparsity-promoting regularizers $\psi(\coefm)$.
The simplest choice of such a regularizer is $\psi(\coefm)=\lambda \sum_{i=1}^{\nsamples} \| \coefv_i \|_1$ (with $\lambda>0$),
for which the pursuit problem can be split into $\nsamples$ independent problems on the columns of $\coefm$,
\begin{eqnarray}
\min_{\coefv\in\reals^{\natoms}}
\frac{1}{2}\norm{\datav-\dictm\coefv}_2^2 + \lambda \| \coefv \|_1.
\label{eq:reg-lasso}
\end{eqnarray}
This is the classical \emph{unstructured} sparse coding problem, often referred to as Lasso \citep{tibshirani96} or basis pursuit \citep{CDS99}.

\emph{Structured} sparse models further assume that the pattern of the non-zero coefficients of $\coefm$ exhibits a specific structure known \emph{a priori}. Let $A\subseteq \{1,\ldots,\natoms\}$ denote groups of indices of atoms.
Then, we define a group structure, $\groupset$, as a collection of groups of atoms, $\groupset = \{A_1,\ldots,A_{\ngroups} \}$. The regularizer corresponding to the group structure is defined as the column-wise sum
\begin{equation}
\psi_\groupset(\coefm) = \sum_{\ai=1}^{\nsamples} \psi_\groupset(\coefv_\ai)
\quad \textrm{where} \quad
\psi_\groupset(\coefv) = \sum_{\gi=1}^{\ngroups} \lambda_\gi \| \coefv_\gi \|_2,
\label{eq:reg-group}
\end{equation}
and $\coefv_\gi$ denotes the subvector of $\coefv$ corresponding to the group of atoms $A_\gi$.
The regularizer function $\psi$ in the Lasso problem (\ref{eq:reg-lasso}) arises from
the special case of singleton groups $\groupset=\setdef{\{1\},\{2\},\ldots,\{\natoms\}}$ and setting $\weight_\gi =\weight$.
As such, the effect of $\psi_\groupset$ on the groups of $\coefv$ is a natural generalization of the one obtained with unstructured sparse coding: it ``turns
on'' and ``off'' atoms in groups according to the structure imposed by $\groupset$.

Several important structured sparsity settings can be cast as particular cases of (\ref{eq:reg-group}):
\emph{Group sparse coding}, a generalization of the standard sparse coding to the cases in which the dictionary is sub-divided into groups \citep{yuan06}, in this case $\groupset$ is a partition of $\{1,\ldots,\natoms\}$; 
\emph{Hierarchical sparse coding}, assuming a hierarchical structure of the non-zero coefficients \citep{JournalHiLasso,zhao-2009-37,JenattonMOB11}. The groups in $\groupset$ form a hierarchy with respect to the inclusion relation (a tree structure), that is, if two groups overlap, then one is completely included in the other one;
\emph{Overlapping group sparse coding}, relaxing the hierarchy assumption so that groups of atoms $A_r$ are allowed to overlap. This model was found to be successful in modeling gene expression and other genetic data.  
Note that in all the above cases, the structure is repeated across the columns of $\coefm$; consequently, the structured sparse coding problem (\ref{eq:parsimonious.modeling}) can be split into $\nsamples$ independent problems operating on the columns of $\datam$ and $\coefm$.
One possible way of extending sparse models is by imposing structure on sub-matrices of $\coefm$.
\emph{Collaborative sparse coding} generalizes the concept of structured sparse coding to collections of input vectors by promoting given patterns of non-zero elements in the coefficient matrix \cite{JournalHiLasso,ER10}.

%

%


\subsection{Low-rank models and robust PCA}

Another significant manifestation of parsimony typical to many classes of data is low rank.
The classical low rank model is \emph{principal component analysis} (PCA),
in which the data matrix $\datam \in \reals^{\ndims \times \nsamples}$ (each column of $\datam$ is an $\ndims$-dimensional data vector), is decomposed into $\datam = \bb{L} + \bb{E}$,
where $\bb{L}$ is a low rank matrix and $\bb{E}$ is a perturbation matrix.
%

PCA is known to produce very good results when the perturbation is small \citep{pca}. However, its performance is highly sensitive to the presence of samples not following the model; even a single outlier in the data matrix $\datam$ can render the estimation of the low rank component arbitrarily far from the true matrix $\bb{L}$.
\cite{Candes2011-JACM,xu2012robust} proposed to robustify the model by adding a new term to the decomposition to account for the presence of outliers,
$\datam = \bb{L} + \bb{O} + \bb{E}$, where $\bb{O}$ is an outlier matrix with a sparse number of non-zero coefficients of arbitrarily large magnitude.
In one of its formulations, the \emph{robust principal component analysis} can be pursued by solving the convex program
\begin{eqnarray}
\min_{\bb{L},\bb{O}\in\reals^{\ndims \times \nsamples}}
\frac{1}{2}\norm{\datam-\bb{L}-\bb{O}}_\mathrm{F}^2 + \lambda_\ast \| \bb{L} \|_\ast + \lambda \|\bb{O} \|_1.
\label{eq:rpca}
\end{eqnarray}
The same way the $\ell_1$ norm is the convex surrogate of the $\ell_0$ norm (i.e., the convex norm closest to $\ell_0$),
the nuclear norm, denoted as $||\cdot ||_\ast$, is the convex surrogate of matrix rank. 
%
The parameter $\lambda_\ast$ controls the tradeoff between the data fitting error and the rank of the approximation.

In \cite{srebro2005rank} it was shown that the nuclear norm of a matrix of $\bb{L}$ can be reformulated as a penalty over all possible factorizations
\begin{equation}
\norm{\bb{L}}_\ast =  \min_{\bb{A}, \bb{B}}
\frac12 \norm{\bb{A}}_{\mathrm{F}}^2 + \frac12 \norm{\bb{B}}_{\mathrm{F}}^2 \quad \quad \textrm{s.t.} \quad \bb{A}\bb{B} = \bb{L},
\label{ec.nuclear.norm.eq}
\end{equation}
The minimum is achieved through the SVD of $\bb{L} = \bb{U} \bb{\Sigma}\bb{V}^T$: the minimizer of \eqref{ec.nuclear.norm.eq} is
$\bb{A}=\bb{U}\bb{\Sigma}^{\frac12}$ and $\bb{B}=\bb{\Sigma}^{\frac12}\bb{V}$. This factorization has been recently exploited in
parallel processing across multiple processors to produce state-of-the-art algorithms for matrix completion problems \citep{Recht:2011wv},
as well as an alternative approach to robustifying PCA in \cite{mateos-2011}.

In \eqref{eq:rpca}, neither the rank of $\bb{L}$ nor the level of sparsity in $\bb{O}$ are assumed known \emph{a priori}.
However, in many applications, it is a reasonable to have a rough upper bound of the rank, say $\textrm{rank}(\bb{L})\leq q$.
Combining this with \eqref{ec.nuclear.norm.eq}, it was proposed in \cite{mateos-2011} to reformulate \eqref{eq:rpca} as
\begin{equation}
\min_{\dictm_0, \bb{S}, \bb{O}}
\frac{1}{2} \norm{\datam - \dictm_0\bb{S} - \bb{O}}^2_{\mathrm{F}} + \frac{\lambda_\ast}{2} ( \norm{\dictm_0}_{\mathrm{F}}^2 + \norm{\bb{S}}_{\mathrm{F}}^2 ) + \lambda \norm{\bb{O}}_1,
\label{ec.convex.unconstrained.US}
\end{equation}
with $\dictm_0\in\reals^{\ndims \times q}$, $\bb{S}\in\reals^{q\times \nsamples}$, and $\bb{O}\in\reals^{\ndims \times \nsamples}$.
This new factorized formulation reduces the number of optimization variables and reveals much structure hidden in the problem.
The low rank component can now be thought of as an under-complete dictionary
$\dictm_0$, with $q$ atoms, multiplied by a matrix $\bb{S}$ containing in its columns the corresponding coefficients for each data vector in $\datam$.
This interpretation allows to write problem \eqref{eq:rpca} in the form of our general parsimonious models \eqref{eq:parsimonious.modeling},
with $\coefm = (\bb{S}; \bb{O})$, $\psi(\coefm) = \frac{\lambda_\ast}{2} \norm{\bb{S}}_{\mathrm{F}}^2  + \lambda \norm{\bb{O}}_1$,
$\dictm=(\dictm_0, \bb{I}_{\ndims \times \ndims})$, and $\phi(\dictm) = \frac{\lambda_\ast}{2} \norm{\dictm_0}_{\mathrm{F}}^2$. 
Furthermore, unlike the nuclear norm, this new formulation
of the rank-reducing regularization is differentiable and vector-wise separable (well suited for the online setting \eqref{eq:parsimonious.modeling.online}).
However, problem \eqref{ec.convex.unconstrained.US} is no longer convex. Fortunately, it can be shown that any stationary point of \eqref{ec.convex.unconstrained.US}, $\{\dictm_0,\bb{S},\bb{O}\}$, satisfying $||\datam - \dictm_0\bb{S} - \bb{O}||_2\leq \lambda_\ast$ is an globally optimal solution of \eqref{ec.convex.unconstrained.US} \citep{morteza}.
Thus, problem \eqref{ec.convex.unconstrained.US} can be solved using an alternating minimization, as in our online setting,
without the risk of falling into a stationary point that is not globally optimal.

\subsection{Non-negative matrix factorization}

Another popular low rank model is non-negative matrix factorization (NMF).
Given a non-negative data matrix $\datam \in \reals^{\ndims \times \nsamples}$, NMF aims at finding a factorization
$\datam \approx \dictm\coefm$ into non-negative matrices $\dictm \in \reals^{\ndims \times q}$ and $\coefm \in \reals^{q \times \nsamples}$, with $q\leq n$.
The factorization is obtained by solving the highly non-convex problem
\begin{equation}
\min_{\dictm, \coefm \ge \bb{0}}
\norm{\datam - \dictm\coefm}^2_{\mathrm{F}}.
\label{eq:nmf}
\end{equation}
Problem \eqref{eq:nmf} can be stated as particular instance of (\ref{eq:parsimonious.modeling}) by setting $\psi$ and $\phi$ to be the sum of element-wise indicator functions of the form
\begin{eqnarray}
i^+(t) &=& \left\{ \begin{array}{ll}
                                   0 & : ~ t \ge 0 \\
                                   \infty & : ~ t < 0.
                                 \end{array}
 \right.
 \end{eqnarray}
The non-negativity constrain has been shown to be crucial for learning a part representation of the data, making it particularly
attractive in the problem of source separation.
An extensive amount of work reported in the literature has been devoted to regularizing \eqref{eq:nmf} in meaningful ways.

Similarly to PCA, NMF is sensitive to outliers in the data matrix. A robust variant can be obtained by adding a sparse outlier term to the decomposition, $\datam \approx \dictm_0\bb{S} + \bb{O}$, as done in the RPCA model \cite{zhang2011robust}.
Again here the problem can be cast as a particular instance of \eqref{eq:parsimonious.modeling} defining $\coefm = (\bb{S}; \bb{O})$ and $\dictm=(\dictm_0, \bb{I}_{\ndims \times \ndims})$.

NMF is by construction a low rank representation, since the number of atoms in $\dictm_0$ is normally chosen to be significantly smaller
than the dimension of the data. This means that an upper bound of the rank of the approximation needs to be known beforehand.
NMF is known to be very sensitive to this parameter, due to the natural compromise between richness of the model and over-fitting. 
In most practical settings, $q$ is carefully chosen based on empirical evidence.
In \cite{SBS_ismir}, we proposed a cure to this phenomenon by incorporating a rank-reducing term into (\ref{eq:nmf}), establishing in this way a link between NMF and the RPCA problem \eqref{ec.convex.unconstrained.US}. Combined with the outlier term, we can formulate a robust low-rank NMF problem
\begin{equation}
\min_{\dictm_0, \bb{S}, \bb{O} \ge \bb{0}}
\norm{\datam - \dictm_0\bb{S} - \bb{O} }^2_{\mathrm{F}} + \lambda_\ast \norm{\dictm_0\bb{S} }_\ast + \lambda \norm{\bb{O}}_1.
\label{eq:rnmf}
\end{equation}
The resemblance of (\ref{eq:rnmf}) to the RPCA problem tempts to apply the reasoning we used before to get rid of the nuclear norm, adding non-negativity constraints
to (\ref{ec.convex.unconstrained.US}),
\begin{equation}
\min_{\dictm_0, \bb{S}, \bb{O} \ge \bb{0}}
\frac{1}{2} \norm{\datam - \dictm_0\bb{S} - \bb{O}}^2_{\mathrm{F}} + \frac{\lambda_\ast}{2} ( \norm{\dictm_0}_{\mathrm{F}}^2 + \norm{\bb{S}}_{\mathrm{F}}^2 ) + \lambda \norm{\bb{O}}_1.
\label{eq:rnmf:US}
\end{equation}
However, unlike the RPCA case, problems (\ref{eq:rnmf}) and (\ref{eq:rnmf:US}) are not equivalent as the minimum of (\ref{ec.nuclear.norm.eq}) is not necessarily
attained by non-negative factors.
In fact, adding non-negativity constraints to (\ref{ec.nuclear.norm.eq}) produces
\begin{eqnarray}
|| \dictm_0\bb{S} ||_\ast & \leq &  \frac12\min_{\mat{A},\mat{B} \ge \mat{0} }  \left\{ ||\mat{A} ||^2_\mathrm{F} +||\mat{B} ||^2_\mathrm{F}
\,\, \textrm{s.t.} \,\, \bb{A}\bb{B} = \dictm_0\bb{S} \right\} \nonumber\\
&\leq& \frac12 \norm{\dictm_0}_\mathrm{F}^2 + \frac12 \norm{\bb{S}}_\mathrm{F}^2.
 \label{eq.nucelar3}
 \end{eqnarray}
Thus, the sum of the Frobenius norms of the non-negative matrices $\dictm$ and $\bb{S}$ gives an upper bound on the nuclear norm of their product.
While not being fully equivalent to \eqref{eq:rnmf}, the objective in problem \eqref{eq:rnmf:US} still achieves both robustness to outliers and rank regularization. With some abuse of terminology, we will refer to problem \eqref{eq:rnmf:US} as to RNMF.
Again here, \eqref{eq:rnmf:US} is a particular instance of \eqref{eq:parsimonious.modeling} well suited for the online setting.

\section{Proximal methods}
\label{sec:prox}

Proximal splitting is a powerful optimization technique allowing to efficiently solve a variety of optimization problems, such as non-smooth convex programs. They have been adopted by the machine learning and signal processing communities for their simplicity, convergence guarantees, and the fact that they are well suited for tackling sparse and structured sparse coding problems that can be written as (\ref{eq:parsimonious.modeling}) (refer to \cite{Bach11} for recent reviews). In Section~\ref{sec:learning} we will use these algorithms to construct
efficient learnable pursuit processes.

Proximal splitting methods are designed for solving optimization problems in which the cost function can be split into the sum of two terms,
one convex and differentiable with an $\alpha$-Lipschitz continuous gradient, and another convex extended real valued and possibly non-smooth.
Clearly, pursuit problems \eqref{eq:parsimonious.modeling.online} fall into this category: the convex quadratic data fitting term has a linear gradient $\dictm^\Tr (\dictm \coefv - \datav)$
with the Lipschitz constant given by the squared spectral norm of the dictionary, $\alpha = \norm{\dictm}^2$, and the regularizer $\psi$ is typically convex and non-smooth.
The proximal splitting method with fixed constant step defines a series of iterates,
$\{\coefv^{k}\}_{k\in \naturals}$,
\begin{equation}
\coefv^{k+1} = \prox_{\alpha \psi}(\coefv^{k} - \frac{1}{\alpha} \dictm^\Tr (\dictm \coefv^k - \datav) ),
\label{ec.iterates}
\end{equation}
where
\begin{equation}
\prox_{\alpha \psi}(\coefv) = \argmin{\mat{u}\in \reals^\ndims} \, || \mat{u} - \coefv||_2^2 + \alpha \psi(\mat{u})
\end{equation}
 denotes the proximal operator of $\psi$.
%
%
%
Fixed-step algorithms have been shown to have relatively slow sub-linear convergence, and
many alternatives have been studied in the literature to improve the convergence rate \cite{fista,nesterov07}.
Accelerated versions of the fixed-step algorithm can be used to reach linear convergence rates (the best possible for the class of first order methods).
The discussion of theses methods is beyond of the scope of this paper.


Proximal splitting methods become particularly interesting when the proximal operator of $\psi$ can be computed exactly and efficiently.
Many important cases of structured sparsity fall into this category.
For the simple unstructured sparsity models induced by regularizers of the form
$$\psi(\bb{\coefv}) =  \norm{\bb{\lambda} \odot \bb{\coefv}}_1 = \sum_{i=1}^\ndims \lambda_i z_i,$$
with $\odot$ denoting element-wise multiplication,
the proximal operator reduces to the element-wise scalar soft-thresholding operator,
$(\bb{\pi}_{\bb{\lambda}}(\coefv))_i =  \tau_{\lambda_i} (z_i)$, with $\tau_\lambda(t) = \sgn(t)\max \{0,|t|-\lambda \}$.
In this case, the fixed step proximal splitting algorithm corresponds
to the popular iterative shrinkage-thresholding algorithm (ISTA)
\cite{daubechies2004iterative,fista} summarized in Algorithm~\ref{alg:ISTA}.
Note that the matrices $\bb{H}$ and $\bb{W}$ in Algorithm~\ref{alg:ISTA}
are derived from the linear gradient of the data term.

\begin{algorithm}[t]

\SetKwInOut{Input}{input}\SetKwInOut{Output}{output}

\Input{Data $\datav$, dictionary $\dictm$, weights $\bb{\lambda}$.}

\Output{Sparse code $\coefv$.}

Define $\bb{H} = \bb{I} - \frac{1}{\alpha}  \dictm^\Tr \dictm$, $\bb{W} = \frac1\alpha \dictm^\Tr$, $\bb{t} =  \frac{1}{\alpha}\bb{\lambda}$.

Initialize $\coefv^0 = \bb{0}$ and
$\bb{b}^0 = \bb{W} \datav$.

\For{$k=1,2,\dots$ until convergence}
{


$\coefv^{k+1} = \prox_{\bb{t}}(\bb{b}^{k})$

$\bb{b}^{k+1} = \bb{b}^{k} + \bb{H} (\coefv^{k+1} - \coefv^{k}$)

}
\caption{Iterative shrinkage-thresholding algorithm (ISTA).  \label{alg:ISTA} }

\end{algorithm}
\begin{algorithm}[tb!]

\SetKwInOut{Input}{input}\SetKwInOut{Output}{output}

\Input{Data $\bb{x}$, dictionary $\bb{D}_0$, weights $\lambda, \lambda_\ast$.}

\Output{Approximation $\bb{l}$, outlier $\bb{o}$.}

Define $\bb{H} = \bb{I} - \frac{1}{\alpha}\left(
                                            \begin{array}{cc}
                                              \dictm_0^\Tr \dictm_0 + \lambda_\ast\bb{I} &  \dictm_0^\Tr \\
                                              \dictm_0 & (1+\lambda_\ast)\bb{I} \\
                                            \end{array}
                                          \right)$,
$\bb{W} = \frac{1}{\alpha}\left(
                                            \begin{array}{c}
                                              \dictm_0^\Tr\\
                                              \bb{I} \\
                                            \end{array}
                                          \right)$, and $\bb{t} = \frac{\lambda}{\alpha} \left(
                                            \begin{array}{c} \bb{0}\\ \bb{1}\\ \end{array}
                                          \right)$.

Initialize $\mat{z}^0 = \bb{0}$, $\bb{b}^0 = \bb{W} \bb{x}$.

\For{$k=1,2,\dots$ until convergence}
{

$\bb{z}^{k+1} = \prox_{\bb{t}}( \bb{b}^k )$

$\bb{b}^{k+1} = \bb{b}^k + \bb{H} (\bb{z}^{k+1} - \bb{z}^k)$

}

Split $\bb{z}^{k+1} = (\bb{s}; \bb{o})$ and output $\bb{l} = \dictm_0\bb{s}$.

\caption{Proximal descent algorithm for the online RPCA and RNMF problems with fixed dictionary $\dictm_0$.
The distinction between the two models is obtained through the selection of the proximal operator:
$\prox_{\bb{t}}(\bb{b}) = \tau_{\bb{t}}(\bb{b})$ (RPCA) and
$\prox_{\bb{t}}(\bb{b}) = \tau^+_{\bb{t}}(\bb{b})$ (RNMF).
\label{alg:RPCA-RNMF}}

\end{algorithm}

If $\bb{z}$ is furthermore constrained to be non-negative, as the sparse outlier in the RNMF problem, the soft thresholding is replaced by its one-sided counterpart
$\tau^+_\lambda(t) = \max \{0,t-\lambda \}$. The proximal operator of the indicator function $i^+(t)$ imposing the non-negativity constraint
is simply $\tau^+_0(t)=\max \{0,t\}$.
The fixed-step proximal descent algorithm for the online RPCA and RNMF problems is summarized in Algorithm~\ref{alg:RPCA-RNMF}.

To generalize the proximal operator to the structured sparsity case,
let us first consider an individual term in the sum (\ref{eq:reg-group}), $\psi_\gi(\coefv) = \lambda_\gi \norm{\coefv_\gi}_2$.
Its proximal operator, henceforth denoted as $\bb{\pi}_{\lambda_\gi}$, can be computed as,
\begin{equation}
(\bb{\pi}_{\lambda_\gi}(\coefv))_s= \left\{
\begin{array}{lcl}
\displaystyle{\frac{\coefv_\gi}{\norm{\coefv_\gi}_2} \tau^+_{\lambda_\gi}(\norm{\coefv_\gi}_2) } & :~ s= \gi,\\
\coefv_\gi& : ~s \neq \gi,\\
\end{array}
\right.
\label{ec.reg.single}
\end{equation}
where sub-indices $r$ and $s$ denote a sub-vector of the $\natoms_\gi$-dimensional vector specified by the group of atoms $A_\gi$.

Note that $\bb{\pi}_{\lambda_\gi}$ applies a group soft thresholding to the coefficients belonging to the $\gi$-th group and leaves the remaining ones unaffected.
For a non-overlapping collection of groups $\groupset$, the
proximal operator of $\psi_\groupset$ is group-separable and can be computed independently for each group as
$$\bb{\pi}^\groupset_{\bb{\lambda}}(\coefv)= ( (\bb{\pi}_{\lambda_1}(\coefv))_1 ; \cdots ; (\bb{\pi}_{\lambda_{|\groupset|}}(\coefv))_{|\groupset|} ),$$
where $\bb{\lambda} = (\lambda_1, \dots, \lambda_{|\groupset|})^\Tr$
is the vector with the threshold parameters $\lambda_\gi$. 


In general, when the groups of $\groupset$ overlap, there is no efficient way of computing the proximal operator of $\psi_\groupset$.
An important exception to this is the hierarchical setting with tree-structured groups.
%
Let us be given a tree hierarchy of groups $\groupset = \groupset_1 \cup \cdots \cup \groupset_L$ with each $\groupset_l$ aggregating $r_l$ non-overlapping
groups corresponding to the $l$-th level of the tree.
Then, the proximal operator of $\psi_\groupset$ can be shown to be given by the composition of the proximal operators of
$\psi_{\groupset_l}$ in ascending order from the leaves to the root \cite{JenattonMOB11,JournalHiLasso},
$\bb{\pi}^\groupset = \bb{\pi}^{\groupset_1}_{\bb{\lambda}_1} \circ \cdots \circ \bb{\pi}^{\groupset_L}_{\bb{\lambda}_L}$.
Here $\bb{\lambda}_l \in \mathbb{R}^{r_l}$ denotes the sets of weights corresponding to the constituent group of each level.
A particular case of the tree-structured hierarchical sparse model is the two-level HiLasso model
introduced to simultaneously promote sparsity at both group and coefficient level \cite{JournalHiLasso,friedman10a}.
Algorithm~\ref{alg:ISTA} is straightforward to generalize to the case of hierarchical sparsity by using the appropriate proximal operator.

%

It is worthwhile noting that the update in Algorithm~\ref{alg:ISTA} can be applied to a single element (or group in case of structured sparsity) at a time in a 
(block) coordinate manner. Several variants of coordinate descent (CoD) and block-coordinate descent (BCoD) proximal methods have been proposed \cite{Tseng,Osher}. 
Typically, one proceeds as in Algorithm~\ref{alg:ISTA}, first applying the proximal operator 
$\bb{y} =\bb{\pi}(\bb{b}^{k})$. Next, the residual $\bb{e} = \bb{y} - \coefv^{k}$ is evaluated, and the group is selected e.g.
according to $\gi = \mathrm{arg}\max_\gi \norm{\bb{e}_\gi }_2$ (in case of unstructured sparsity, $\gi = \mathrm{arg}\max_\gi |\bb{e}_\gi|$).
Then, $\bb{b}^{k+1}$ is computed by applying $\bb{H}$ only to the selected subgroup of $\bb{e}$, and $\coefv^{k+1}$ is computed by replacing
the subgroup of $\coefv^k$ with the corresponding subgroup of $\bb{y}$.
%

%
%
%
%
%
%
%
%
%
%
%
%
%
%

\section{Learnable pursuit processes}
\label{sec:learning}

The general parsimonious modeling problem (\ref{eq:parsimonious.modeling}) can be alternatively viewed as the minimization problem
\begin{eqnarray}
\mathop{\min_{\enc : \mathbb{R}^\ndims \rightarrow \mathbb{R}^\natoms}}_{\dec : \mathbb{R}^\natoms \rightarrow \mathbb{R}^\ndims}
\frac{1}{\nsamples}  \sum_{i=1}^{\nsamples}  L(\datav_i,\enc,\dec),
\label{eq:parsimonious.modeling.new}
\end{eqnarray}
with $L(\datav,\enc,\dec) = \frac{1}{2}\norm{ (\bi{id} - \dec \circ \enc)(\datav)  }_\mathrm{F}^2 +\psi(\enc (\datav)) + \phi(\dec)$.
The optimization is now performed over an \emph{encoder} $\coefv = \enc(\datav)$ mapping the data vector $\datav$
to the representation vector $\coefv$, and a \emph{decoder} $\datav = \dec(\coefv)$ performing the converse mapping.
The encoder/decoder pair is sought to make the composition $\dec \circ \enc$ close to the identity map, under the regularity constraints
promoted by the penalties $\psi$ and $\phi$.

Existing parsimonious models restrict the decoder to the class of linear functions $\dec_\dictm(\coefv) = \dictm \coefv$ parametrized
by the dictionary $\dictm$.
For a fixed dictionary $\dictm$, the optimal encoder is given by
\begin{eqnarray}
\enc^\ast & = & \mathrm{arg}\min_{\enc : \mathbb{R}^\ndims \rightarrow \mathbb{R}^\natoms} L(\datav,\enc,\dec_{\dictm}),
\label{eq:encoders.modeling.opt-enc}
\end{eqnarray}
which is nothing but the solution of the representation pursuit problem obtained through
and iterative optimization algorithm, such as the proximal methods described in Section~\ref{sec:prox}.
This interpretation is possible since the solution of the pursuit problem implicitly defines a deterministic mapping that assigns to each input vector $\datav\in\reals^n$ a unique parsimonious code $\coefv\in \reals^m$. Naturally, this mapping cannot be stated explicitly.

%
%

In contrast, the process-centric approach proposed in this work, aims at formulating a modeling scheme were both the encoder and the decoder can be explicitly stated and efficiently computed. In our proposed framework, the encoders are constructed explicitly as parametric deterministic functions, $\enc_{\bb{\Theta}}: \reals^n\rightarrow \reals^m$ with a set of parameters collectively denoted as $\bb{\Theta}$, while the decoders are the exact same simple linear decoders, $\dec_\dictm(\coefv) = \dictm \coefv$, used in model-centric approaches (we relax this assumption in the following section).
We denote by $\F$ the family of the parametric functions $\enc_{\bb{\Theta}}$.
Naturally, two fundamental question arise: how to select the family $\F$ capable of defining good parsimonious models,
and how to efficiently select the best parameters $\bb{\Theta}$ given a specific family.
We will refer to the first problem as to selecting the \emph{architecture} of the pursuit process, while the second will be referred to as \emph{process learning}.
We start with the latter, deferring the architecture selection to Section~\ref{sec:architectures}.

\subsection{Process learning}
\label{sec:process.learning}

With the process-centric perspective in mind, problem (\ref{eq:parsimonious.modeling.new}) can be stated as
\begin{eqnarray}
\min_{\enc_{\bb{\Theta}} \in \F, \dictm \in \reals^{\ndims \times \natoms} }
\frac{1}{\nsamples}  \sum_{i=1}^{\nsamples}  L(\datav_i,\enc_{\bb{\Theta}},\dec_\dictm),
\end{eqnarray}
where the family $\F$ imposes some desired characteristics on the encoder such as continuity and almost everywhere differentiability, and
certain computational complexity.
As in \eqref{eq:parsimonious.modeling}, this problem can be naturally solved using an alternating minimization scheme,
sequentially minimizing for $\enc_{\bb{\Theta}}$ or $\dictm$ while leaving the other one fixed.
Note that when the encoder $\enc_{\bb{\Theta}}$ is fixed, the problem in $\dictm$ remains essentially the same dictionary update problem and can be solved
exactly as before.
In what follows, we therefore concentrate on solving the process learning problem
\begin{eqnarray}
\min_{\enc_{\bb{\Theta}} \in \F}
\frac{1}{\nsamples}  \sum_{i=1}^{\nsamples}  L(\datav_i,\enc_{\bb{\Theta}}),
\label{eq:proc-learning}
\end{eqnarray}
given a fixed dictionary. To simplify notation, we henceforth omit $\dec_\dictm$ from $L$ whenever $\dictm$ is fixed.

Observe that problem (\ref{eq:proc-learning}) attempts to find a process $\enc_{\bb{\Theta}}$ in the family $\F$
that minimizes the empirical risk over a finite set of training examples, as an approximation to the expected
risk
\begin{eqnarray*}
\hat{\L}(\enc_{\bb{\Theta}}) = \frac{1}{\nsamples}  \sum_{i=1}^{\nsamples}  L(\datav_i,\enc_{\bb{\Theta}})  \approx \int L(\datav,\enc_{\bb{\Theta}}) dP(\datav) = \L(\enc_{\bb{\Theta}})
\end{eqnarray*}
over the data distribution $P$.
While the empirical risk measures the encoder performance over the training set,
the expected risk measures the expected performance over new data samples following the
same distribution, that is, the generalization capabilities of the model.
When the family $\F$ is sufficiently restrictive, the statistical learning theory justifies minimizing the empirical risk instead of the
expected risk\cite{Vapnik71}. We will come back to this issue in Section~\ref{sec:approximation.accuracy}, where we address the accuracy of the proposed
encoders in approximating $\L(\enc_{\bb{\Theta}})$. 

When the functions belonging to $\F$ are almost everywhere differentiable with respect to the parameters $\bb{\Theta}$,
stochastic gradient descent (SGD) can be used to optimize \eqref{eq:proc-learning}, with almost sure convergence to a stationary point \cite{bottou-2010}.
%
At each iteration, a random subset of the training data, $\{\datav_{i_1},\ldots,\datav_{i_r}\}$,
is selected and used to produce an estimate of the (sub)-gradient of the objective function.
Specifically, in our case the parameters of $\coefv_{\bb{\Theta}}$ are updated as
\begin{equation}
\bb{\Theta} \leftarrow \bb{\Theta} - \mu \frac{1}{r} \sum_{k=1}^{r}  \frac{\partial  L(\datav_{i_k},\enc_{\bb{\Theta}}) }{\partial \bb{\Theta}},
\label{eq:encoders.sgd}
\end{equation}
where $\mu$ is a decaying step, repeating the process until convergence.
This requires the computation of the (sub)-gradients $\partial L / \partial\bb{\Theta}$,
which is achieved by a back-propagation procedure as detailed in the sequel.
SGD algorithms scale well to big data applications where the limiting factor
is the computational time rather than the number of available samples.

\subsection{Approximation accuracy}
\label{sec:approximation.accuracy}

Following \cite{bottou-2010}, we split the process training approximation error into three terms,
$\epsilon = \epsilon_{\textrm{app}} + \epsilon_{\textrm{est}} + \epsilon_{\textrm{opt}}$.
The \emph{approximation error}
$\epsilon_{\textrm{app}} = \mathbb{E} \{ \L(\enc^\ast_{\bb{\Theta}}) - \L(\enc^\ast)\}$
measures how well the optimal unrestricted pursuit process $\enc^\ast$ given by \eqref{eq:encoders.modeling.opt-enc}
is approximated by the optimal pursuit process restricted to $\F$,
$\enc^\ast_{\bb{\Theta}} = \mathrm{arg} \min_{ \enc_{\bb{\Theta}} \in \F } \L(\enc_{\bb{\Theta}})$.
The \emph{estimation error}
$\epsilon_{\textrm{est}} = \mathbb{E} \{ \L(\hat{\enc}^\ast_{\bb{\Theta}}) - \L(\enc^\ast_{\bb{\Theta}})\}$
with $\hat{\enc}^\ast_{\bb{\Theta}} = \mathrm{arg} \min_{ \enc_{\bb{\Theta}} \in \F } \hat{\L}(\enc_{\bb{\Theta}})$
measures the cost of optimizing the empirical risk instead of the expected risk.
Finally, the \emph{optimization error}
$\epsilon_{\textrm{opt}} = \mathbb{E} \{ \L(\hat{\enc}_{\bb{\Theta}}) - \L(\hat{\enc}^\ast_{\bb{\Theta}})\}$
measures the effect of having $\hat{\enc}_{\bb{\Theta}}$ that
minimizes the empirical risk only approximately.

The estimation error vanishes asymptotically with the increase of the training set size $n$.
The optimization error can be made negligible (at least, in the offline setting) by simply increasing the number of SGD iterations.
Consequently, the quality of the learned pursuit process largely depends on the choice of the family $\F$, which we will address in the next section.


\subsection{Process architecture}
\label{sec:architectures}

We extend the ideas introduced in \cite{LecunNN} to derive families of trainable pursuit processes from proximal methods.
Let us examine a generic fixed-step proximal descent algorithm described in Section~\ref{sec:prox}.
%
%
Each iteration can be described as a function receiving the current state $(\bb{b}_{\mathrm{in}}, \bb{z}_{\mathrm{in}})$ and
producing the next state $(\bb{b}_{\mathrm{out}}, \bb{z}_{\mathrm{out}})$ by applying the non-linear transformation
$\coefv_{\mathrm{out}} = \prox_{\bb{t}}(\bb{b}_{\mathrm{in}})$ (representing the proximal operator),
and the linear transformation $\bb{b}_{\mathrm{out}} = \bb{b}_{\mathrm{in}} + \bb{H} (\coefv_{\mathrm{out}} - \coefv_{\mathrm{in}})$ (representing the linear part of the gradient).
This can be described by the function $(\bb{b}_{\mathrm{out}}, \bb{z}_{\mathrm{out}}) = \bi{f}_{\bb{H},\bb{t}}(\bb{b}_{\mathrm{in}}, \bb{z}_{\mathrm{in}})$ parametrized by the matrix $\bb{H}$ describing the linear transformation, and the vector $\bb{t}$ describing the parameters of the proximal operator $\prox_{\bb{t}}$.

A generic fixed-step proximal descent algorithm can therefore be expressed as a long concatenation of such iterations,
$
\enc^\ast(\bb{x}) = \cdots \circ \bi{f}_{\bb{H},\bb{t}} \circ \cdots \circ \bi{f}_{\bb{H},\bb{t}} (\bb{W} \datav, \bb{0}),
$
with the initialization $(\bb{b}_{\mathrm{in}}, \bb{z}_{\mathrm{in}}) = (\bb{W} \datav, \bb{0})$.
For example, Algorithms~\ref{alg:ISTA} and \ref{alg:RPCA-RNMF} follow this structure exactly, for the appropriate choice of the parameters.
Block-coordinate proximal descent algorithms operate very similarly except that the result of the application of $\bi{f}$ is substituted to
a subset of the elements of the state vector, selected in a state-dependent manner at each iteration.

Following \cite{LecunNN}, we consider the family of pursuit processes derived from \emph{truncated} proximal descent algorithms with $T$ iterations,
$\F_T = \{ \enc_{T,\bb{\Theta}}(\datav) = \bi{f}_{\bb{H},\bb{t}} \circ \cdots \circ \bi{f}_{\bb{H},\bb{t}} (\bb{W} \datav, \bb{0}) \}$.
For convenience, we collected all the process parameters into a single pseudo-vector $\bb{\Theta} = \{ \bb{W},\bb{H}, \bb{t} \}$.
Also note that at the last iteration, only $\bb{z}$ of the state vector $(\bb{b},\bb{z})$ is retained; with some abuse of notation, we still denote the last iteration by $\bi{f}$.
Finally, we denote by $\F_\infty$ the family of untruncated processes.

\begin{figure*}
\begin{center}

\includegraphics[width=0.6\columnwidth]{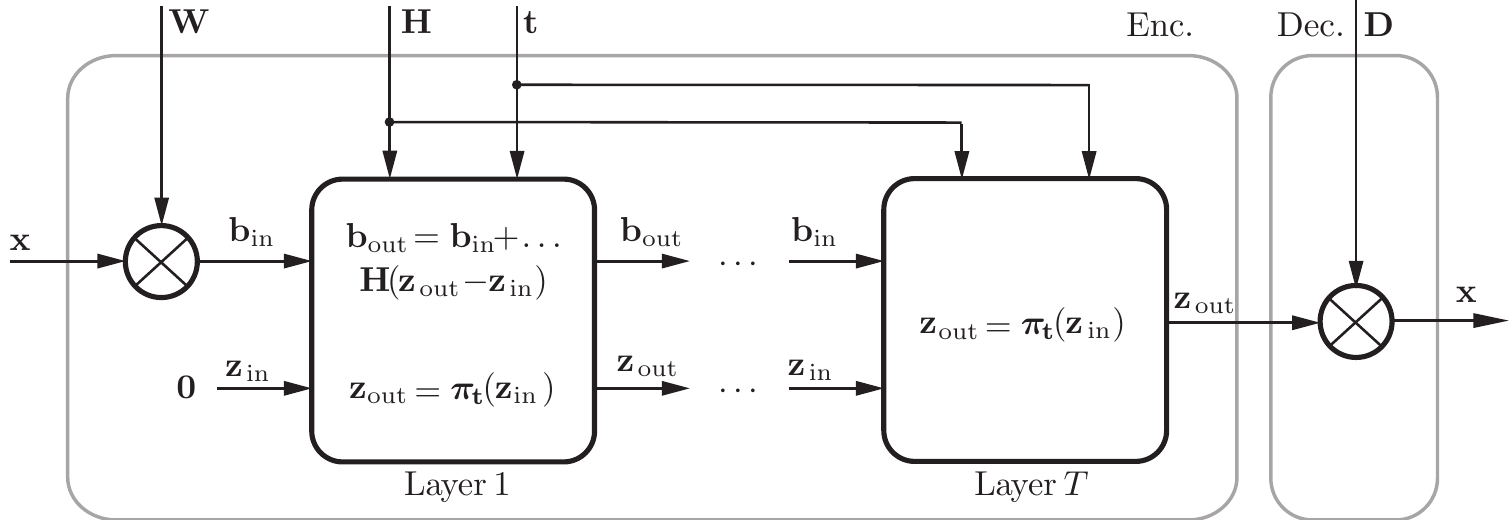}\\[1ex]
Unstructured Lasso encoder architecture\\[3ex]
\includegraphics[width=0.6\columnwidth]{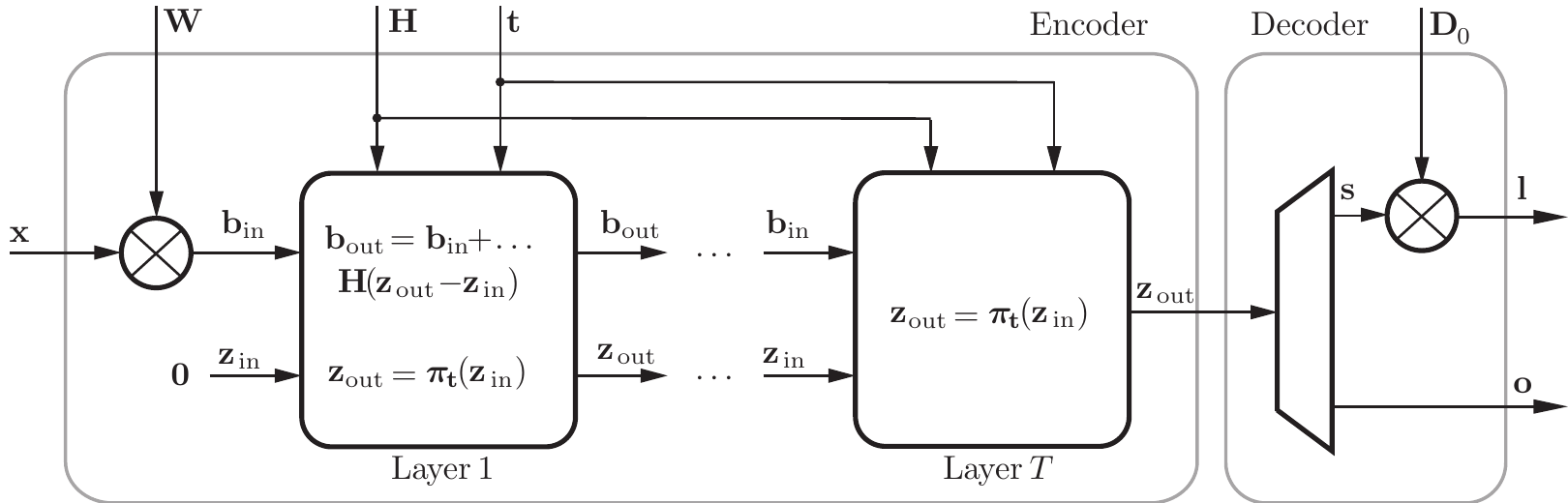}\\[1ex]
RPCA/RNMF encoder architecture
\end{center}
\caption{\label{fig:architecture}  Encoder/decoder architectures for the unstructured Lasso 
(top),
and the robust PCA/NMF 
(bottom).}
\end{figure*}

A process $\enc_{T,\bb{\Theta}} \in \F_T$ can be thought of as a feed-forward neural network with identical layers $\bi{f}_{\bb{H},\bb{t}}$.
Flow diagrams of processes derived from the proximal descent Algorithms~\ref{alg:ISTA} and \ref{alg:RPCA-RNMF} for the Lasso and RPCA/RNMF problems are
depicted in Figure~\ref{fig:architecture}.
Since the processes are almost everywhere $\mathcal{C}^1$ with respect to the input $\datav$ and the parameters $\bb{\Theta}$,
it is possible to calculate the sub-gradients required for the optimization algorithms.
The computation of the sub-gradients of $L (\datav, \enc_{\bb{\Theta}}(\datav))$ with respect to $\bb{\Theta}$ is carried out by an iterated application of the chain rule starting at the output and propagating backward into the network.
%
The procedure, frequently referred to as back-propagation, is detailed in Algorithm~\ref{alg:bprop}, where following the standard notation from the neural network literature, the $\delta$ prefix denotes the gradient of $L$ with respect to the variable following it, $\delta \ast = \partial L / \partial \ast$.

\begin{algorithm}[tb]

\SetKwInOut{Input}{input}\SetKwInOut{Output}{output}

\Input{Sub-gradient $\displaystyle{\delta \bb{z} = \frac{\partial L(\bb{z}^T)}{\partial \bb{z}}}$.}

\Output{Sub-gradients of $L$ with respect to the parameters, $\delta \bb{H}$, $\delta \bb{W}$, $\delta \bb{t}$; and with respect to the
input, $\delta \bb{x}$.}



Initialize $\displaystyle{\delta \bb{t}^T = \frac{\partial \bb{\pi}(\bb{b}^T)}{\partial \bb{t}} \delta \bb{z}}$, $\displaystyle{\delta \bb{b}^T = \frac{\partial \bb{\pi}(\bb{b}^T)}{\partial \bb{b}} \delta \bb{z}}$, $\delta \bb{H}^T = \bb{0}$, and $\delta \bb{z}^T = \bb{0}$

\For{$k=T-1,T-2,\dots,1$}
{

$\displaystyle{\delta \bb{H}^{k-1} = \delta \bb{H}^{k} + \delta \bb{b}^k (\bb{z}^{k+1} - \bb{z}^k)^\Tr}$

$\displaystyle{\delta \bb{t}^{k-1} = \delta \bb{t}^k + \frac{\partial \bb{\pi}(\bb{b}^k)}{\partial \bb{t}}(\bb{H}^\Tr \delta\bb{b}^k - \delta\bb{z}^k )}$

$\displaystyle{\delta \bb{z}^{k-1} = -\bb{H}^\Tr \delta \bb{b}^k}$

$\displaystyle{\delta \bb{b}^{k-1} = \delta \bb{b}^k + \frac{\partial \bb{\pi}(\bb{b}^k)}{\partial \bb{t}}\bb{H}^\Tr \delta\bb{b}^k + \frac{\partial \bb{\pi}(\bb{b}^k)}{\partial \bb{b}} \delta\bb{z}^k }$

}

Output $\delta \bb{H} = \delta \bb{H}^0$, $\delta \bb{t} = \delta \bb{t}^0$,
$\displaystyle{\delta \bb{W} = \delta \bb{b}^k \bb{x}^\Tr}$, and $\displaystyle{\delta \bb{x} = \bb{W}^\Tr \delta \bb{b}^k}$

\caption{Computation of the sub-gradients of $L(\datav,\enc_{T,\bb{\Theta}}(\datav))$ for a pursuit process $\enc_{T,\bb{\Theta}} \in \F_T$.
The reader is warned not to confuse the $T$-th iteration index, $\bb{z}^T$, with the transpose, $\bb{z}^\Tr$. \label{alg:bprop} }

\end{algorithm}

\subsection{Approximation error vs. complexity trade-off}
\label{sec:approx.error}

Since the objective minimized via proximal descent is convex, it is guaranteed that there exists some selection of the parameters $\bb{\Theta}^\ast$ such that
$\lim_{T \rightarrow \infty } \enc_{T,\bb{\Theta}^\ast} = \enc^\ast$.
In other words, the optimal process $\enc^\ast$ is contained in $\F_\infty$.
Furthermore, reformulating the non-asymptotic convergence analysis from \cite{fista} in our language, the following holds:
\begin{theorem}[Beck\&Teboulle]
For every $\dictm$, there exists $\bb{\Theta}^\ast$ and $C > 0$ such that for every $\coefv$ and $T \ge 1$,
$\displaystyle{L(\bb{x},\enc_{T,\bb{\Theta}^\ast}) - L(\bb{x},\enc^\ast) \le \frac{C}{2T} \| \enc^\ast(\datav)  \|_2^2}$.
\end{theorem}

\noindent This result is worst-case, in the sense that it holds for every input vector $\datav$. Assuming bounded support of the input distribution and taking expectation with respect to it, we obtain the following:
\begin{corollary}
There exist $\bb{\Theta}^\ast$ and $C>0$ such that for $T \ge 1$, 
$$\displaystyle{\epsilon_{\mathrm{app}} = \mathbb{E} \{ \L(\enc_{T,\bb{\Theta}^\ast}) - \L(\enc^\ast) \} \le \frac{C}{2T}}.$$
\end{corollary}

\noindent In other words, the family $\F_T$ of pursuit processes allows to set the approximation error to an arbitrarily small number.

One may wonder whether the iter we have undergone so far is of any value at all, given that the process learning approach is only capable of \emph{approximating} the optimal pursuit process achieved via an iterative algorithm.
This picture totally changes, however, when we consider the trade-off between the approximation error, $\epsilon_{\mathrm{app}}$, and the computational complexity
of the encoder, which is proportional to $T$.
The bound in Theorem~1 is uniform for every input $\datav$, independently of the input distribution $P$, while in practice we would strive for a much faster decrease of the error on more probable inputs. This implies that the bound in Corollary~1 is by no means tight, and there might be
other selections of $\bb{\Theta}$ giving much lower approximation error at a fixed $T$. While pursuit processes optimal in the sense of the expected risk can be found using the process learning approach, there is no simple way for an iterative pursuit algorithm to take the input data distribution into account.

\begin{figure}
\begin{center}
\includegraphics[width=0.49\columnwidth]{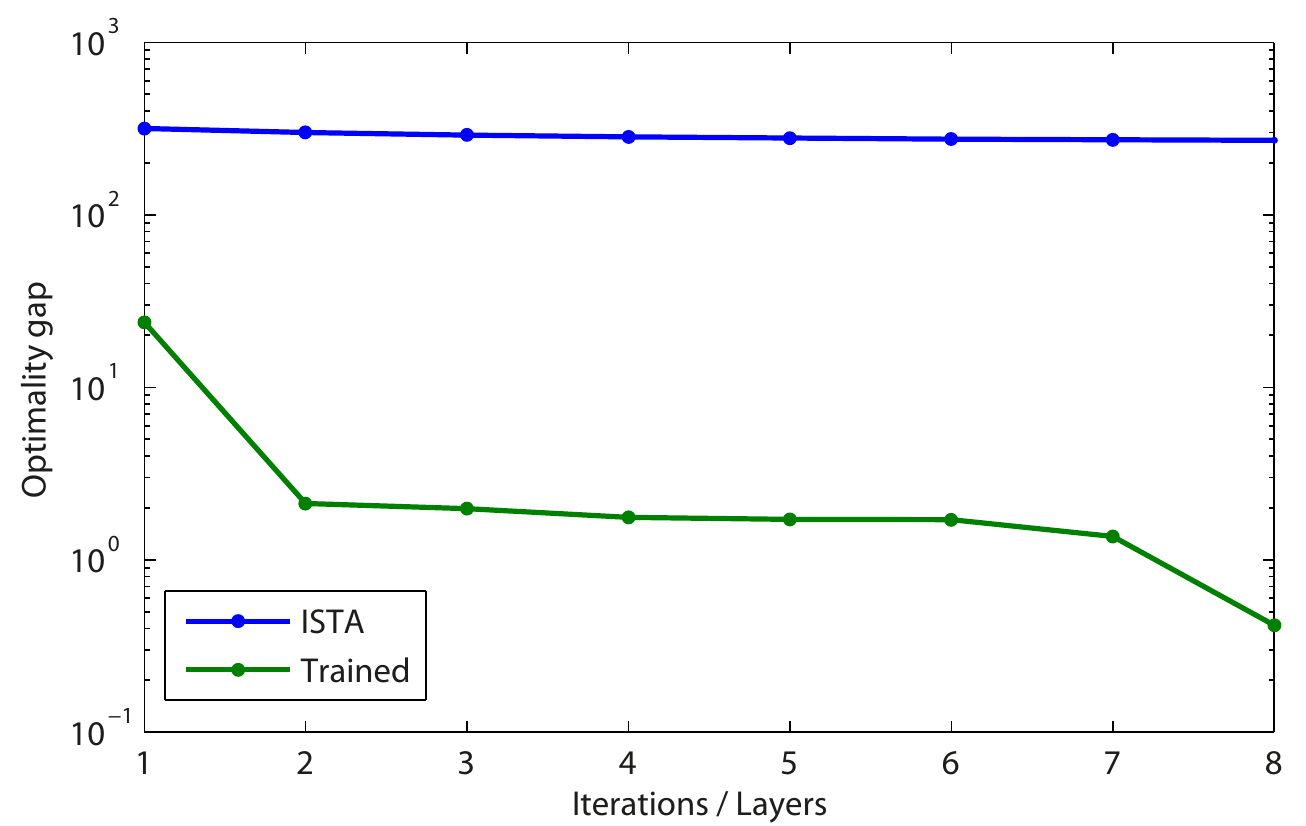}
\includegraphics[width=0.49\columnwidth]{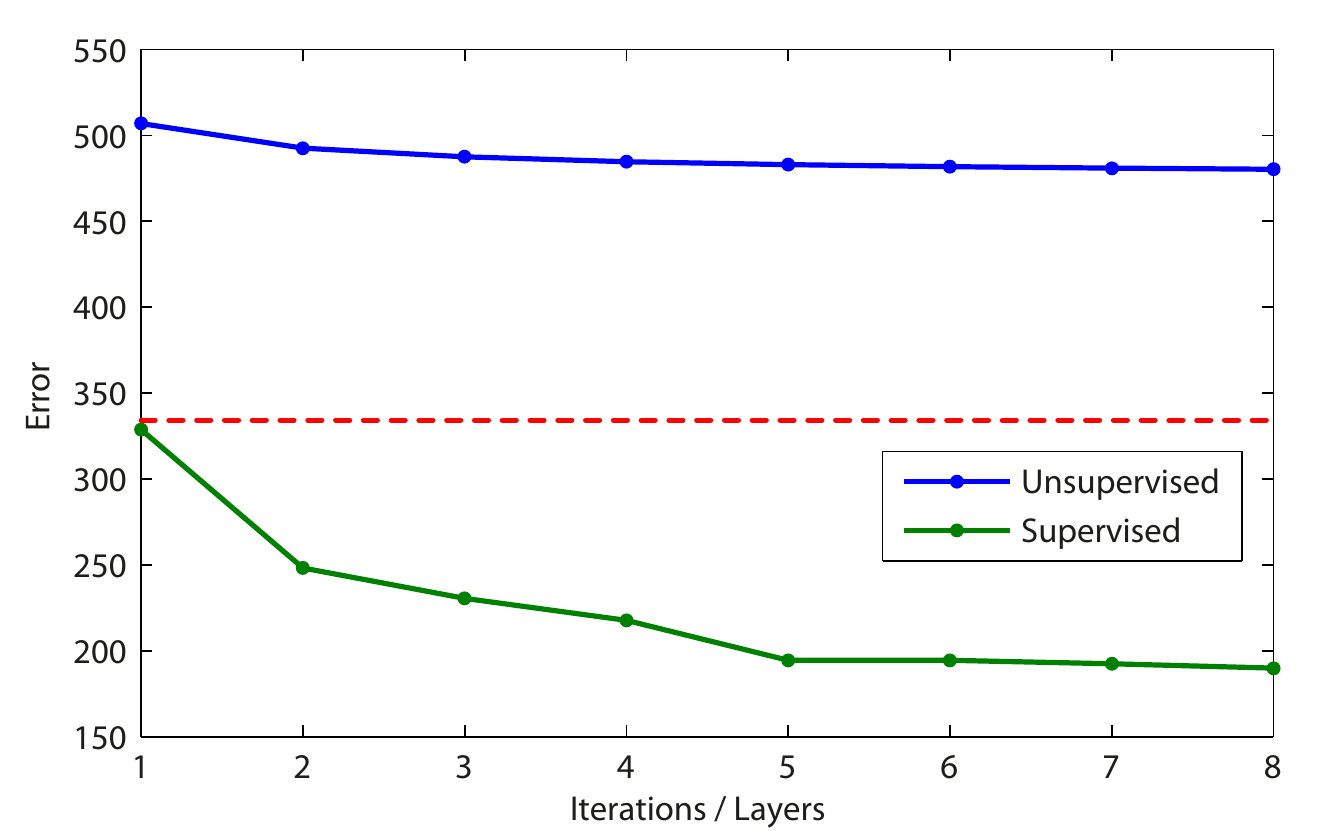}
\end{center}
\caption{\label{fig:approx.error} Comparison between RNMF encoders and proximal methods for music source separation.
Optimality gap as a function of the number of iterations/layers with encoders trained in the unsupervised regime (left);
and the $\ell_2$ separation error obtained with supervised and the unsupervised training (bottom). The red dotted line represents
the error achieved by Algorithm~\ref{alg:RPCA-RNMF} after convergence. }
\end{figure}

An illustration to the advantage of the trained pursuit process
is shown in Figure~\ref{fig:approx.error} (left), in which the performance of RNMF encoders is compared to that of the corresponding proximal algorithms.
As the example, we used the audio separation problem described in further details in sections~\ref{sec:approx.error} and \ref{sec.experiments}.
The figure shows the optimality gap $\hat{\L}(\enc_{T,\bb{\Theta}}) - \hat{\L}(\enc^\ast)$ as a function of $T$
for the truncated proximal descent ($\Theta=\Theta^0$ set as prescribed by Algorithm~\ref{alg:RPCA-RNMF}) and for the trained encoder ($\Theta = \Theta^\ast$).
This optimality gap can be thought of as an empirical approximation error, $\epsilon_{\mathrm{app}}$, for the corresponding spaces $\F_T$.
It takes about $70$ iterations of the proximal method to reach the error obtained by a $7-$layer encoder.
A further and stronger justification to the process-centric approach advocated in this paper is presented in the next section.

\section{Training regimes}
\label{sec:training.regimes}

As it was described in the previous section, parsimonious modeling
can be interpreted as training an encoder/decoder pair $(\enc,\dec)$ that minimizes the empirical loss
$\hat{\L}$ in problem (\ref{eq:parsimonious.modeling.new}).
We refer to this setting as \emph{unsupervised}, as no information beside the data samples themselves is provided at training.
An important characteristic of this training regime
is that it can be performed online, combining online adaptation of $\bb{\Theta}$
with online dictionary update as described in Section~\ref{sec.sparse.models}.
%

In many applications, one would like to train the parsimonious model comprising the encoder/decoder pair to
optimally perform a specific task, e.g., source separation, object recognition, or classification.
This setting can be still viewed as the solution of the modeling problem (\ref{eq:parsimonious.modeling.new}), with the modification
of the training objective $L$ to include the task-specific knowledge.
However, for most objectives, given a fixed linear decoder $\dec_{\dictm}(\coefv) = \dictm \coefv$, the encoder of the form
\begin{eqnarray}
\enc(\datav) & = & \mathrm{arg}\min_{ \coefv \in \mathbb{R}^\natoms }
\frac{1}{2}\norm{ \datav - \dictm \coefv  }_\mathrm{F}^2 + \psi(\coefv)
\label{eq:pursuit-enc}
\end{eqnarray}
is no more optimal, in the sense that it is not the solution of (\ref{eq:parsimonious.modeling.new}) with the fixed decoder.
Moreover, the inclusion of an encoder of the form (\ref{eq:pursuit-enc}) into the modeling problem (\ref{eq:parsimonious.modeling.new})
gives rise to the hard \emph{bi-level} optimization problem
\begin{eqnarray}
\min_{\dictm \in \mathbb{R}^{\ndims \times \natoms}}
\frac{1}{\nsamples}  \sum_{i=1}^{\nsamples}  L(\datav_i,\coefv_i,\dec_{\dictm})
\quad \mathrm{s.t.} ~ \coefv_i = \mathrm{arg}\min_{ \coefv \in \mathbb{R}^\natoms }
\frac{1}{2}\norm{ \datav_i - \dictm \coefv  }_\mathrm{F}^2 + \psi(\coefv).
\end{eqnarray}

Trainable pursuit processes provide an advantageous alternative in these cases. First, by being an explicit deterministic function, they remove the need to solve the bi-level optimization problem. Second, they have fixed complexity and latency controllable through the parameter $T$, and they are expected to achieve a better complexity-approximation error trade-off than the iterative pursuit processes. Third, trainable processes constitute a good compromise between the two extremes: the iterative pursuit having no elements of learning at all, and the general regression problems, fully relying on learning.
The described trainable processes have the structure of the iterative pursuit built into the architecture on one hand, while
leaving tunable degrees of freedom that can improve their modeling capabilities on the other. Furthermore, such processes come with a very good initialization of the parameters, which is non-trivial in the more general learning scenarios.

In the sequel, we exemplify various training regimes (i.e., different objectives in (\ref{eq:parsimonious.modeling.new})) that can be used in combination with the architectures described in Section~\ref{sec:architectures}, leading to a comprehensive framework of process-centric parsimonious models.

\subsection{Supervised learning}

In many applications, parsimonious models are used as a first order approximation to various classes of natural signals.
An example is the music separation problem discussed in Section~\ref{sec:approx.error}.
While achieving excellent separation results, both the RPCA and RNMF models are merely a crude representation of the reality: the music is never exactly low-rank, as well as the voice does not exactly admit the na\"{\i}ve unstructured sparse model.
We propose to fill the gap between the parsimonious models and the more sophisticated (and, hence, prone to errors) domain-specific models,
by incorporating learning.
This can be done by designing the objective function $L$ of (\ref{eq:parsimonious.modeling.new}) to incorporate the domain-specific information.
Figure~\ref{fig:approx.error} (bottom) shows that encoders trained this way outperform not only their unsupervisedly trained counterparts, but also the exact RPCA/RNMF algorithm.

Sticking to the example of audio source separation by means of online robust PCA or NMF,
let us be given a collection of $n$ data vector, $\datav_i$, representing the short-time spectrum of the
mixtures, for which the clean ground-truth accompaniment $\bb{l}_i^\ast$ and voice $\bb{o}_i^\ast$ spectra are given.
We aim at finding an RPCA/RNMF encoder $(\bb{s};\bb{o}) = \enc_{\bb{\Theta}}(\datav)$ with the architecture depicted in Figure~\ref{fig:architecture},
and the linear decoder $(\bb{l},\bb{o}) = \dictm (\bb{s}; \bb{o})$ parametrized by $\dictm=(\dictm_0,\bb{I}_{\ndims \times \ndims})$, that minimize (\ref{eq:parsimonious.modeling.new})
with the objective
\begin{eqnarray}
\lefteqn{\hat{\L}(\bb{z}_{\bb{\Theta}},\dictm) = \frac{1}{n} \sum_{i=1}^n \| \dictm_0\bi{s}_{\bb{\Theta}}(\datav_i) - \bb{l}_i^\ast  \|_2^2 + \| \bi{o}_{\bb{\Theta}}(\datav_i) - \bb{o}_i^\ast  \|_2^2 } \nonumber\\
&& + \frac{\lambda_\ast}{2}\left(  \| \dictm_0 \|_2^2 + \| \bi{s}_{\bb{\Theta}}(\datav_i) \|_{\mathrm{F}}^2 \right) +  \lambda \| \bi{o}_{\bb{\Theta}}(\datav_i) \|_1. \hspace{1.5cm}
\label{eq:obj-sup}
\end{eqnarray}
Note that the architecture and the initial parameters of the encoder are already a very good starting point, yet the supervised training allows it to better model
the reality that is not fully captured by the RPCA/RNMF model (Figure~\ref{fig:approx.error}, bottom).

In a broader perspective, the sound separation example can be viewed as a particular instance of \emph{supervised} encoder/decoder training regimes, in which the loss $L$ is expressed in terms of the composite output of $\dec \circ \enc$, but is supervised by some $\bb{y}_i$ different from the input data, e.g.,
\begin{equation}
\hat{\L}(\enc,\dec) = \frac{1}{n} \sum_{i=1}^n \| \bb{y}_i - \dec \circ \enc (\datav_i)  \|_2^2 + \psi(\enc(\datav_i) ) + \phi(\dec).
\label{eq:train-sup}
\end{equation}
Compare this to the unsupervised setting, where essentially $\bb{y}_i = \datav_i$, and the performance of the encoder/decoder pair is measured by their ability to \emph{reconstruct} the input.

The idea of \cite{LecunNN} to train pursuit processes to approximate the output of iterative pursuit algorithms falls into this category.
For each training data vector $\datav_i$, let $\coefv_i^\ast = \enc^\ast(\datav_i)$ be the output of the optimal encoder (\ref{eq:encoders.modeling.opt-enc}).
Setting the decoder to the identity map, $\dec = \bi{id}$ in (\ref{eq:train-sup}) reduces the supervision to the encoder outputs,
\begin{equation}
\hat{\L}(\enc) = \frac{1}{n} \sum_{i=1}^n \| \coefv_i - \enc (\datav_i)  \|_2^2.
\label{eq:train-approx}
\end{equation}
While producing good results in practice, under the limiting factor that the testing data is from the same class as the training data, this training regime requires supervision yet it is unable to improve the modeling capabilities of the encoder beyond those of the underlying parsimonious model (unlike the supervision of the decoder outputs in (\ref{eq:train-sup})). On the other hand, we show in Section~\ref{sec.experiments} that similar performance can be achieved by using unsupervised training. We refer to this regime as \emph{Approximation}.

\subsection{Discriminative learning}
\label{sec:training.regimes.discriminative}

The supervised training setting is also useful to extend parsimonious models beyond the conventional \emph{generative} scenario, in which the data can be approximately recovered from the representation, to the \emph{discriminative} scenario, such as classification problems, where the representation is typically non-invertible and is intended to capture various invariant properties of the data.

As an illustration, we use the simultaneous speech denoising and speaker identification model from \cite{denoising-2012}, in which the spectrogram of the input signal is decomposed into $\datam \approx \bb{D}_0 \bb{S} + \bb{D}\bb{O}$, where $\bb{D}_0 \bb{S}$ capturing the noise is required to be low-rank, while the activation $\bb{O}$ representing the speech is required to be sparse.
A collection of speaker-specific dictionaries, $\bb{D}_1,\dots,\bb{D}_k$, is trained offline, and the models are fit to previously unobserved data. The lowest fitting error is then used to assign the speaker identity. See Section~\ref{sec:robust.speaker.identification} for experimental evaluation with real data.

Using the process-centric methodology, we construct $k$ encoders $(\bi{s}_{\bb{\Theta}_j},\bi{o}_{\bb{\Theta}_j})(\datav)$, and $k$ corresponding decoders
$(\dictm_0\bb{s},\dictm_j \bb{o})$ with the shared noise dictionary $\dictm_0$. The encoder/decoder pairs are trained by minimizing an empirical loss of the form
\begin{eqnarray}
L(\datav,l,\Theta_1,\dots,\Theta_k,\dictm_0,\dots,\dictm_k) = & \!\!\!\!\!\!\!\!\!\!\!\| \datav - \bb{D}_0 \bi{s}_{\bb{\Theta}_l} (\datav) - \bb{D}_{l} \bi{o}_{\bb{\Theta}_l} (\datav)  \|_2^2   \label{eq:train-discrim} \\
&\hspace{-18ex}+ \sum_{j\ne l} \max\left\{ 0, \epsilon -  \| \datav - \bb{D}_0 \bi{s}_{\bb{\Theta}_j} (\datav) - \bb{D}_{j} \bi{o}_{\bb{\Theta}_j} (\datav)  \|_2^2  \right\}, \nonumber
\end{eqnarray}
averaged on the training set containing examples of noisy speech, $\datav$, and the corresponding labels, $l \in \{1,\dots,k\}$ indicating which speaker is present in the sample.
For each training sample, the loss function (\ref{eq:train-discrim})
promotes low fitting error for the encoder from the class coinciding with the ground-truth class,
while favoring high fitting error for the rest of the encoders. The hinge parameter $\epsilon$ counters
excessive influence of the negatives.

In Section~\ref{sec:robust.speaker.identification}, we show empirical evidence that encoder/decoder pairs trained using a discriminative loss perform the classification task better than those trained to produce the best reconstruction.

\subsection{Data transformations}
\label{sec:data.transformations}

Parsimonious models rely on the assumption that the input data vectors
are ``aligned'' with respect to each other. This assumption might be violated in many applications.
A representative example is face modeling via RPCA, where the low dimensional model only holds if the facial images are pixel-wise aligned \cite{Peng2011-PAMI}.
Even small misalignments can break the structure in the data; the representation then quickly degrades as the rank of the low dimensional component increases and the matrix of outliers loses its sparsity.
In \cite{Peng2011-PAMI}, it was proposed to simultaneously align the input vectors and solve RPCA by including the transformation
parameters into the optimization variables. This problem is highly non-convex, yet if a good initialization of the transformation parameters is available,
a solution can be found by solving a sequence of convex optimization problems, each of them being comparable to \eqref{eq:rpca}.

Following \cite{Peng2011-PAMI}, we propose to incorporate the optimization over geometric transformations of the input data into our modeling framework.
We assume that the data are known to be subject to a certain class of parametric transformations
$\T = \{ \bi{g}_{\bb{\alpha}}: \mathbb{R}^\ndims \rightarrow \mathbb{R}^\ndims : \bb{\alpha} \}$.
Given a data matrix $\datam$, we search for its best alignment together with finding the best model by solving
\begin{equation}
\mathop{\min_{\bb{z}_{\bb{\Theta}}, \dec_\dictm}}_{\bb{\alpha}_1,\dots,\bb{\alpha}_n}  \frac{1}{\nsamples} \sum_{i=1}^{\nsamples} L( \bi{g}_{\bb{\alpha}_i}(\datav_i),\enc_{\bb{\Theta}},\dec_\dictm),
\label{eq:encoders.modeling.pursuit.transf}
\end{equation}
jointly optimizing over the encoder $\enc_{\bb{\Theta}}$, possibly the decoder $\dec_\dictm$, and the transformation parameters $\bb{\alpha}_i$
of each of the training vectors.
This approach can be used with any of the training objectives described before.

The optimization problem \eqref{eq:encoders.modeling.pursuit.transf} can be carried out as a simple
extension of our general process training scheme.
Transformation parameters $\bb{\alpha}$ can be treated on par with the encoder parameters $\bb{\Theta}$;
note that since the encoder functions $\enc_{\bb{\Theta}}\in \F$ are by construction almost everywhere differentiable with respect to their input,
we can find the sub-gradients of the composition $\enc_{\bb{\Theta}}(\bi{g}_{\bb{\alpha}}(\datav))$ with respect to $\bb{\alpha}$
by simply applying the chain rule.
This allows to solve \eqref{eq:encoders.modeling.pursuit.transf} using SGD and the back-propagation in Algorithm~\ref{alg:bprop}.

The obtained encoders are conceptually very similar to the ones we had before and can still be trained in an online manner.
In this particular setting, our framework presents a significant advantage over existing approaches based on iterative pursuit.
In general, as the new data arrive and the model is refined or adapted, the alignment of the previously received input vectors has to be adjusted.
In our settings, there is no need to recompute the alignment for the whole dataset from scratch, and it can be
adjusted via online SGD.

Unlike the untransformed model, the pursuit is no more given by an explicit function $\enc_{\bb{\Theta}}$, but requires the solution of a simple optimization problem
in $\bb{\alpha}$, $\displaystyle{\bb{\alpha}^\ast =  \mathrm{arg} \min_{\bb{\alpha}} L( \datav,\enc_{\bb{\Theta}} \circ \bi{g}_{\bb{\alpha}},\dec_\dictm)}$.
A new data vector $\bb{x}$ is then encoded as
$\coefv = \enc_{\bb{\Theta}}( \bi{g}_{\bb{\alpha}^\ast}(\datav) )$.

\section{Experimental results}
\label{sec.experiments}

In what follows, we describe experiments conducted to assess the efficiency of the proposed framework.
Fast encoders were implemented in Matlab with built-in GPU acceleration and executed on Intel Xeon E5620 CPU and NVIDIA Tesla C2070 GPU.
When referring to a particular encoder, we specify its architecture and the training regime; for example ``CoD (Unsupervised)''
stands for the CoD network trained in the unsupervised regime.
We denote by \emph{Approximation}, \emph{Supervised}, \emph{Unsupervised} and \emph{Discriminative} the corresponding
training regime as defined in Section~\ref{sec:training.regimes}. \emph{Untrained} denotes an untrained encoder, that is, with parameters set as in
the corresponding fixed-step proximal descent algorithm; the performance of such an encoder coincides with that of a truncated proximal descent.
\begin{figure}[t]
\begin{center}
\includegraphics[width=0.6\columnwidth]{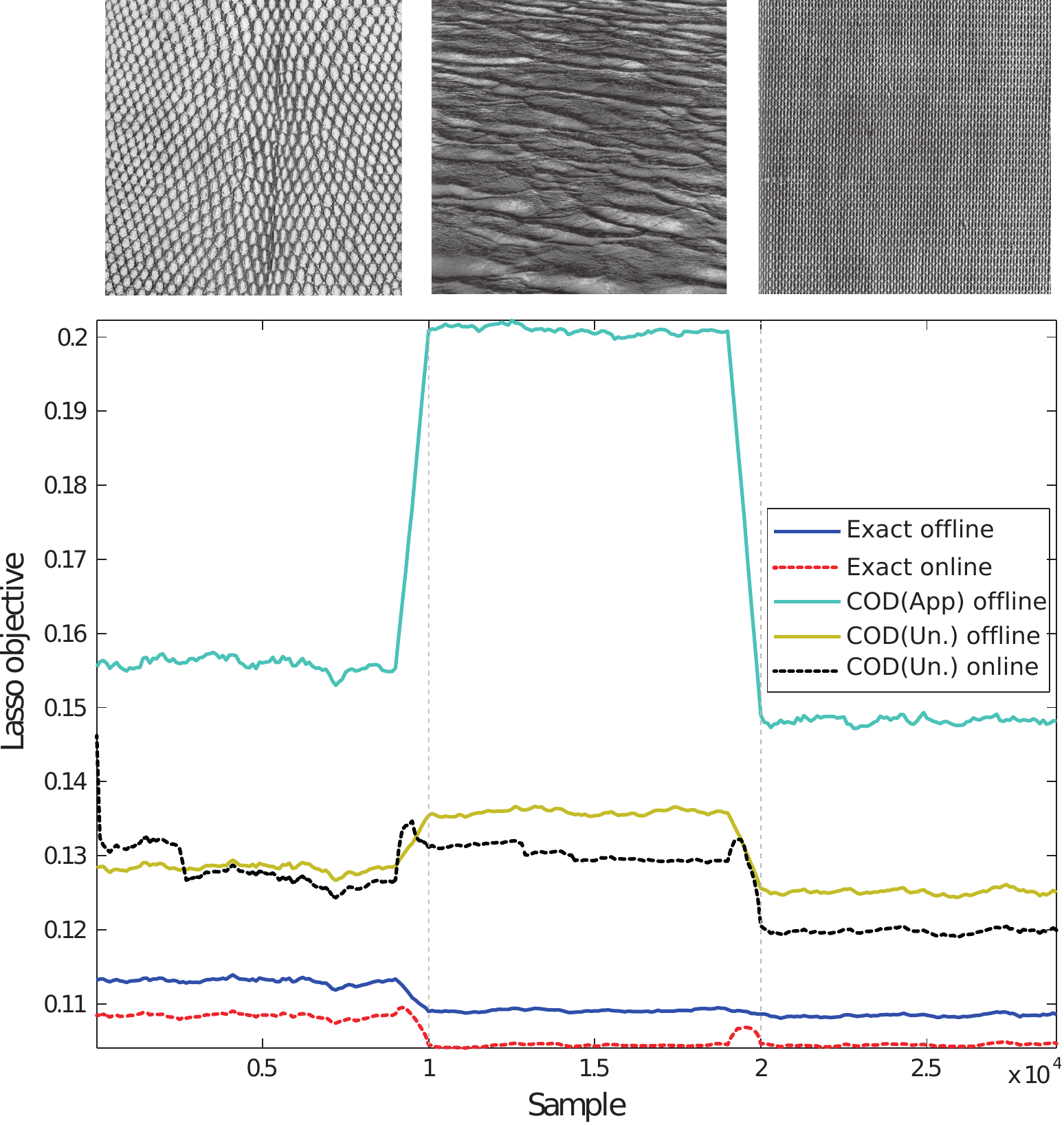}
\end{center}
\caption{Performance of CoD sparse encoders, trained under different training regimes, measured using the Lasso objective, as a function of sample number in the online learning experiment. Shown are the three groups of patches corresponding to different texture images from the Brodatz dataset. \label{fig:online} }
\end{figure}
\subsection{Online sparse encoders}

To evaluate the performance of the unstructured CoD (\emph{Unsupervised}) encoders in the online learning regime, we used $30\times 10^4$ randomly located $8 \times 8$ patches from three images from the Brodatz texture dataset \cite{brodatz}. The patches were
ordered in three consecutive blocks of $10^4$ patches from each image. Dictionary size was fixed to $\natoms = 64$ atoms, and $T=4$ layers were used in all CoD encoders.
Unsupervised online learning was performed using the Lasso objective with $\lambda = 1$ on overlapping windows of $1,\!000$ vectors with a step of $100$ vectors. Online trained encoders were compared to an online version of Lasso with the dictionary adapted on the same data.
As reference, we trained offline a CoD (\emph{Approximation}) encoder and another CoD (\emph{Unsupervised}) encoder. Offline training was performed on a distinct set of $6,\!000$ patches extracted from the same images. 

Performance measured in terms of the Lasso objective is reported in Figure~\ref{fig:online} (high error corresponds to low performance).
%
Initially, the performance of the online trained CoD (\emph{Unsupervised}) encoder is slightly inferior to the offline trained counterpart; however, the online version starts performing better after the network parameters and the dictionary adapt to the current class of data.
The CoD (\emph{Approximation}) encoder trained offline exhibits the lowest performance.
This experiment shows that, while the drop in performance compared to the exact Lasso is relatively low, the computational complexity of the online
CoD (\emph{Unsupervised}) encoder is tremendously lower and fixed.
\begin{table}
\caption{Speaker identification misclassification rates. \label{tab:audio} }
\begin{center}
\begin{tabular}{lcc}
  \hline\hline
\textbf{Code} & \textbf{Error rate} \\
\hline
Exact HiLasso      & 2.35\% \\
NN CoD (\emph{Approximation})   & 6.08\% \\
NN BCoD (\emph{Approximation})  & 3.53\% \\
NN BCoD (\emph{Discriminative}) & 3.44\% \\
  \hline\hline
\end{tabular}
\end{center}
\end{table}
\subsection{Structured sparse encoders}


The performance of the BCoD structured sparse architecture derived from Algorithm~\ref{alg:ISTA} combined with different training regimes is evaluated on a speaker identification task reproduced from \cite{audio}.
In this application the authors use hierarchical sparse coding to automatically detect the speakers in a given mixed signal.
The dataset consists of recordings of five different radio speakers, two females and three males.
$25\%$ of the samples were used for training, and the rest for testing. Within the testing data, two sets of waveforms were created: one containing isolated speakers, and another containing all possible combinations of mixtures of two speakers.
Signals were decomposed into a set of overlapping time frames of $512$ samples with $75\%$ overlap, such that the properties of the signal remain stable within each frame.
An $80$-dimensional feature vector is obtained for each audio frame as its short-time power spectrum envelope (refer to \cite{audio} for details).
Five under-complete dictionaries with $50$ atoms each were trained on the single speaker set minimizing the Lasso objective with  $\lambda = 0.2$ (one dictionary per speaker), and then combined into a single structured dictionary containing $250$ atoms. Increasing the dictionary size exhibited negligible performance benefits.
Speaker identification was performed by first encoding a test vector in the structured dictionary and measuring the $\ell_2$ energy of each of the five groups. Energies were sum-pooled over $500$ time samples selecting the labels of the highest two.

To assess the importance of the process architecture, a CoD (\emph{Approximation}) and a BCoD (\emph{Approximation}) encoders with $T=2$ layers were trained
offline to approximate the solution of the exact HiLasso, with $\lambda_2 = 0.05$ (regularizer parameter in the lower
level of the hierarchy). 
A BCoD (\emph{Discriminative}) encoder with the same settings was also trained in the supervised regime using the discriminative loss function \eqref{eq:train-discrim} to promote or discourage the activation of groups corresponding to knowingly active or silent speakers respectively.

Table~\ref{tab:audio} summarizes the obtained misclassification rates. In agreement with the experiments shown in Section~\ref{sec:architectures}, using an appropriate structured architecture instead of its unstructured counterpart with the same number of layers and the same dictionary increases performance by nearly a factor of two. The use of the discriminative objective further improves performance. Surprisingly, using encoders with only two layers cedes just about $1\%$ of correct classification rate.
The structured architecture showed a crucial roll in producing accurate structured sparse codes. 
Other comparative experiments substantiating this observation can be found in \cite{ICML}.

\subsection{Robust PCA encoders with geometric optimization}


In order to evaluate the performance of robust PCA encoders, we used a face dataset consisting of $800$ $66 \times 48$ images of a female face photographed over the timespan of $4.5$ years, roughly pose- and scale-normalized and aligned.
The images manifested a significant variety of hair and dressing styles, while keeping similar facial traits. Following \cite{Peng2011-PAMI}, we use RPCA to decompose a collection of faces represented as columns of the data matrix $\datam$ into
$\bb{L} + \bb{O}$, with the low-rank term $\bb{L} = \dictm_0 \bb{S}$ approximating the face identity (atoms of $\dictm_0$ can be thought of as ``\emph{eigenfaces}''),
while the outlier term $\bb{O}$ capturing the appearance variability.

\begin{figure*}[t]
\begin{center}
\includegraphics[width=0.85\textwidth]{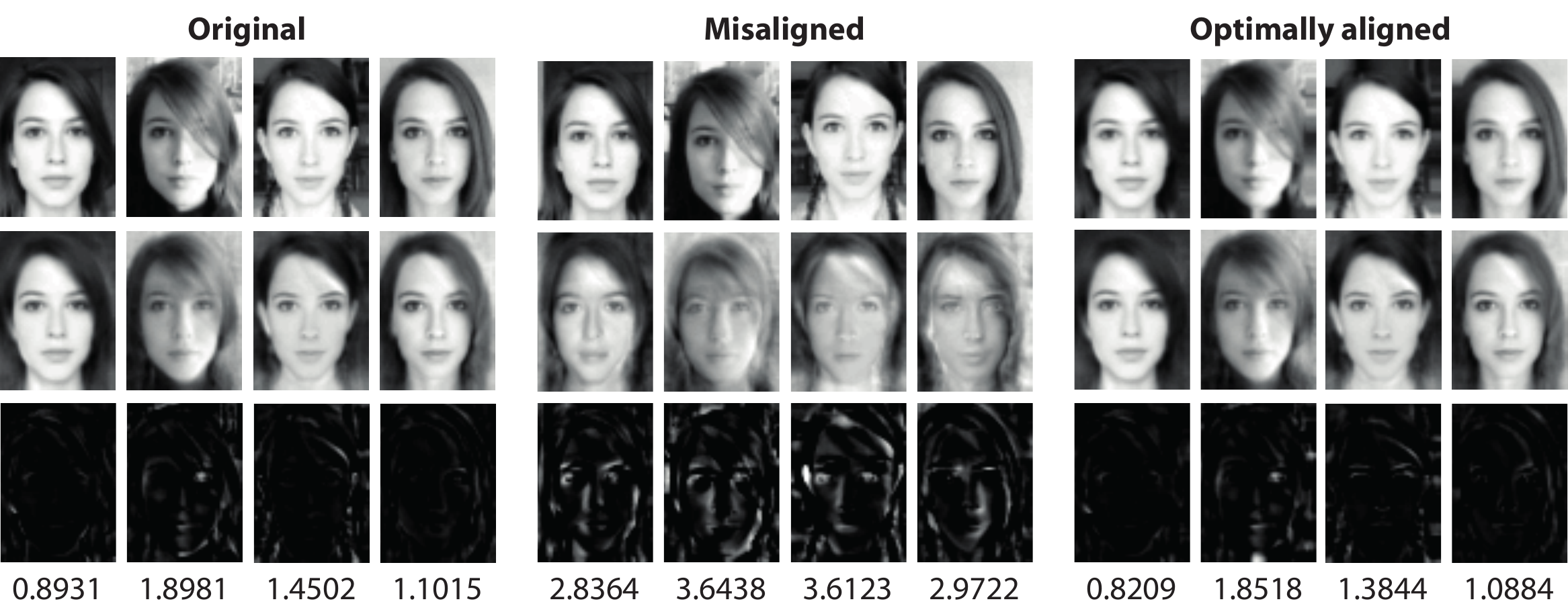}
\end{center}
\caption{Robust PCA representation of the faces dataset in the presence of geometric transformations (misalignment).
Left group: original faces; middle group: shifted faces; right group: faces optimally realigned during encoding.
First row: reconstructed face $\dictm \bb{s}+\bb{o}$; middle row: its low-rank approximation $\dictm \bb{s}$; and bottom row: sparse outlier $\bb{o}$.
RPCA loss is given for each representation.
 \label{fig:geometric} }
\end{figure*}

We trained a five layer RPCA (\emph{Unsupervised}) encoder on $600$ images from the faces dataset.
The dictionary was initialized using standard SVD and parameters were set to $q=50$, $\lambda_\ast=0.1$ and $\lambda = 10^{-2}$.
To evaluate the representation capabilities of RPCA encoders in the presence of geometric transformations,
as a test set, we used the remaining $200$ faces, as well as a collection of geometrically transformed images from the same test set. Sub-pixel planar translations were used for geometric transformations.
The encoder was applied to the misaligned set, optimizing  the unsupervised objective over the transformation parameters.
For reference, the encoder was also applied to the transformed and the untransformed test sets without performing optimization.
Examples of the obtained representations are visualized in Figure~\ref{fig:geometric}.
Note the relatively larger magnitude and the bigger active set of the sparse outlier vector $\bb{o}$ produced for the misaligned faces, and how they are re-aligned when optimization over the transformation is allowed. Since the original data are only approximately aligned, performing optimal alignment during encoding frequently yields lower cost compared to the plain encoding of the original data.
\subsection{Robust NMF encoders}

\subsubsection{Singing voice separation}
\label{sec:voice.separation}

We now evaluate the source separation problem (singing-voice/background-accompaniment), described in Section~\ref{sec:approx.error}.
The separation performance was evaluated on the MIR-1K dataset \cite{hsu2010improvement}, containing $1,\!000$ Chinese karaoke clips performed by amateur singers.
The experimental settings closely followed that of \cite{hsu2010improvement}, to which the reader is referred for further details.
As the evaluation criteria, we used the \emph{normalized source-to-distortion ratio} (NSDR) from the BSS-EVAL metrics \cite{vincent2006performance}, averaged
over the test set.
Encoders with RNMF  architecture composed by $T=20$ layers and $q=25$ were trained using different training regimes.
We used $\lambda = \sqrt{2\nsamples} \sigma$ and $\lambda_\ast  = \sqrt{2}\sigma$ with $\sigma = 0.3$ set following \cite{Candes2011-JACM}.
Table~\ref{tab:comparison} summarizes the obtained separation performance. While (\emph{Unsupervised}) training regime makes fast RNMF encoders on par
with the exact RNMF (at a fraction of the computational complexity and latency of the latter), significant improvement is achieved by using the (\emph{Supervised}) regime,
where the encoders are trained to approximate the ground-truth separation over a reduced training set.
For further details refer to \cite{SBS_ismir}.

\begin {table}[tb]
\caption{Audio separation quality ($dB$ SDR). \label{tab:comparison} }
\begin{center}
\begin{tabular}{lcc}
  \hline\hline
\textbf{Method} &  \textbf{Voice}  &  \textbf{Music} \\
\hline
Ideal frequency masking          & 13.48 & 12.51  \\
Exact RNMF                &	5.60 & 3.17    \\
RNMF (\emph{Untrained})	      & 1.62 & 4.31  \\
RNMF (\emph{Unsupervised})      & 5.00 & 4.01	  \\
RNMF 	(\emph{Supervised})        & 6.36 & 4.32    \\
  \hline\hline
\end{tabular}
\end{center}
\end{table}

%
\subsubsection{Robust speaker identification}
\label{sec:robust.speaker.identification}

The purpose of this experiment is to show the benefits of training parsimonious models in a
discriminative fashion when used for classification tasks.
As an example we use speaker identification in environments heavily contaminated by unstructured noise
described in Section~\ref{sec:training.regimes.discriminative}.
%
We evaluated the classification capabilities of different low rank NMF architectures in combination with two supervised training regimes discussed in
Section~\ref{sec:training.regimes}, one aimed to produce a good
reconstruction of the speech signal and another one optimized to produce the best classification.
The reconstructive approach is analogous to the one used in Section~\ref{sec:voice.separation} for music/singing-voice separation, but using a low rank NMF for both noise and human speech. 
In all our examples we used $T=10$ layers and $q=50$. Parameters $\lambda$ and  $\lambda_\ast$ were chosen as in Section~\ref{sec:voice.separation}.

%

As speech dataset we used a subset of the GRID dataset \cite{cooke2006audio} containing $10$ distinct speakers; each speaker comprising $1,\!000$ short clips.
Three sets of $200$ distinct clips each were used for training, validation, and testing.
%
The GRID clips were artificially contaminated by six categories of noise recorded from different real environments (street, restaurant, car, exhibition, train, and airport) taken from the AURORA corpus \cite{PearceH00}. The voice and the noise clips were mixed linearly with equal energy ($0$ dB SNR).
In all experiments, the spectrogram of each mixture was computed using a window of size $512$ and a step size of $128$ samples (at $8$ KHz sampling rate). 
For further details please refer to \cite{denoising-2012}. Table~\ref{tab:identification} summarizes the classification rates of the compared methods. 
While the results obtained using the low rank NMF (\emph{Supervised}) are very good, they are significantly outperformed by the low rank NMF (\emph{Discriminative}) encoders.
\begin {table}[tb]
\caption{Speaker identification success rate.
  \label{tab:identification} }
\begin{center}
\begin{tabular}{lcccc}
  \hline\hline
\multirow{2}{*}{\textbf{Noise}} & \multirow{2}{*}{\textbf{Exact}}  &   \multicolumn{2}{c}{\textbf{RNMF Encoders}} \\
                                 &                                      &  \textbf{(\emph{Supervised})} & \textbf{(\emph{Discriminative})}  \\
\hline
street	     &	0.86	&	0.91	&	0.91 \\
restaurant	 &	0.91	&	0.89	&	0.90 \\
car	         &	0.90	&	0.91	&	0.96 \\
exhibition	 &	0.93	&	0.91	&	0.95 \\
train	     &	0.93	&	0.88	&	0.96 \\
airport	     &	0.92	&	0.85	&	0.98 \\
\hline
average	     &	0.91	&	0.89	&	0.94 \\
\hline\hline
\end{tabular}
\end{center}
\end{table}

\section{Conclusions and future work}

In this work we have developed a comprehensive framework for process-centric parsimonious modeling.
By combining ideas from convex optimization with multi-layer neural networks,
we have shown how to produce deterministic functions capable of
faithfully approximating the optimization-based solution of parsimonious models at a fraction
of the computational time. Furthermore, at almost the same computational cost, the framework includes different objective functions that allow the encoders to
be trained in a discriminative fashion or solve challenging alignment problems.
We conducted empirical experiments in different settings and real applications such as image modeling, robust face modeling, audio sources separation and  robust speaker recognition. A simple unoptimized implementation already achieves often several order of magnitude speedups when compared to exact solvers.

While we limited our attention to synthesis models, the proposed framework can be naturally extended to  \emph{analysis} cosparse models \cite{nam2012cosparse,vaiter2011robust}, in which the signal is known to be sparse in
a transformed domain. Specifically, given a ``sensing'' matrix $\bb{M}\in \reals^{n\times q}$ and an analysis dictionary $\bb{\Omega}\in  \reals^{p\times m}$, in an analysis counterpart of \eqref{eq:proc-learning}, one looks for a function $\bi{f}\in \F$, where again $\F$ is a space of functions
with certain desired properties, that minimizes
\begin{equation}
\min_{\bi{f}\in\F}  \frac{1}{2} \sum_{i=1}^N \norm{\bb{x}_i - \bb{M} \bi{f}(\bb{x_i})}^2+ \lambda \norm{\bb{\Omega}\bi{f}(\bb{x_i})}_1.
\label{eq:analysis.nn}
\end{equation}
The space $\F$ can be set by truncating suitable iterative optimization algorithms such as the augmented Lagrangian methods of multipliers (ADMM) \cite{bertsekas1999nonlinear}.





\bibliographystyle{IEEEtran}
\bibliography{refs,refs1}

\begin{thebibliography}{10}
\providecommand{\url}[1]{#1}
\csname url@samestyle\endcsname
\providecommand{\newblock}{\relax}
\providecommand{\bibinfo}[2]{#2}
\providecommand{\BIBentrySTDinterwordspacing}{\spaceskip=0pt\relax}
\providecommand{\BIBentryALTinterwordstretchfactor}{4}
\providecommand{\BIBentryALTinterwordspacing}{\spaceskip=\fontdimen2\font plus
\BIBentryALTinterwordstretchfactor\fontdimen3\font minus
  \fontdimen4\font\relax}
\providecommand{\BIBforeignlanguage}[2]{{%
\expandafter\ifx\csname l@#1\endcsname\relax
\typeout{** WARNING: IEEEtran.bst: No hyphenation pattern has been}%
\typeout{** loaded for the language `#1'. Using the pattern for}%
\typeout{** the default language instead.}%
\else
\language=\csname l@#1\endcsname
\fi
#2}}
\providecommand{\BIBdecl}{\relax}
\BIBdecl

\bibitem{CDS99}
S.~Chen, D.~Donoho, and M.~Saunders, ``Atomic decomposition by basis pursuit,''
  \emph{SIAM J. Scientific Computing}, vol.~20, no.~1, pp. 33--61, 1999.

\bibitem{tibshirani96}
R.~Tibshirani, ``Regression shrinkage and selection via the {LASSO},'' \emph{J.
  Royal Stat. Society: Series B}, vol.~58, no.~1, pp. 267--288, 1996.

\bibitem{olshausen1996emergence}
B.~Olshausen and D.~J. Field, ``Emergence of simple-cell receptive field
  properties by learning a sparse code for natural images,'' \emph{Nature},
  vol. 381, no. 6583, pp. 607--609, 1996.

\bibitem{aharon2006img}
M.~Aharon, M.~Elad, and A.~Bruckstein, ``$k$-{SVD}: an algorithm for designing
  overcomplete dictionaries for sparse representation,'' \emph{IEEE Trans. Sig.
  Proc.}, vol.~54, no.~11, pp. 4311--4322, 2006.

\bibitem{yuan06}
M.~Yuan and Y.~Lin, ``Model selection and estimation in regression with grouped
  variables,'' \emph{J. Royal Stat. Society, Series B}, vol.~68, pp. 49--67,
  2006.

\bibitem{jacob2009group}
L.~Jacob, G.~Obozinski, and J.~Vert, ``Group lasso with overlap and graph
  lasso,'' in \emph{ICML}, 2009, pp. 433--440.

\bibitem{zhao-2009-37}
P.~Zhao, G.~Rocha, and B.~Yu, ``The composite absolute penalties family for
  grouped and hierarchical variable selection,'' \emph{Annals of Statistics},
  vol.~37, no.~6A, p. 3468, 2009.

\bibitem{JenattonMOB11}
R.~Jenatton, J.~Mairal, G.~Obozinski, and F.~Bach, ``Proximal methods for
  hierarchical sparse coding,'' \emph{Journal of Machine Learning Research},
  vol.~12, pp. 2297--2334, 2011.

\bibitem{JournalHiLasso}
P.~Sprechmann, I.~Ram\'{\i}rez, G.~Sapiro, and Y.~C. Eldar, ``C-hilasso: A
  collaborative hierarchical sparse modeling framework,'' \emph{IEEE Trans.
  Signal Process.}, vol.~59, no.~9, pp. 4183--4198, 2011.

\bibitem{peleg2012exploiting}
T.~Peleg, Y.~Eldar, and M.~Elad, ``Exploiting statistical dependencies in
  sparse representations for signal recovery,'' \emph{IEEE Trans. Sig. Proc.},
  vol.~60, no.~5, pp. 2286--2303, 2012.

\bibitem{srebro2005rank}
N.~Srebro and A.~Shraibman, ``Rank, trace-norm and max-norm,'' \emph{Proc.
  COLT}, pp. 599--764, 2005.

\bibitem{candes2009exact}
E.~Cand{\`e}s and B.~Recht, ``Exact matrix completion via convex
  optimization,'' \emph{Foundations of Computational mathematics}, vol.~9,
  no.~6, pp. 717--772, 2009.

\bibitem{Candes2011-JACM}
E.~Cand\`{e}s, X.~Li, Y.~Ma, and J.~Wright, ``Robust principal component
  analysis?'' \emph{Journal of the {ACM}}, vol.~58, no.~3, 2011.

\bibitem{xu2012robust}
H.~Xu, C.~Caramanis, and S.~Sanghavi, ``Robust {PCA} via outlier pursuit,''
  \emph{IEEE Trans. Inf. Theory}, vol.~58, no.~5, pp. 3047--3064, 2012.

\bibitem{zhang2011robust}
L.~Zhang, Z.~Chen, M.~Zheng, and X.~He, ``Robust non-negative matrix
  factorization,'' \emph{Frontiers of Electrical and Electronic Engineering in
  China}, vol.~6, no.~2, pp. 192--200, 2011.

\bibitem{NMF}
D.~Lee and H.~Seung, ``Learning parts of objects by non-negative matrix
  factorization,'' \emph{Nature}, vol. 401, no. 6755, pp. 788--791, 1999.

\bibitem{daubechies2004iterative}
I.~Daubechies, M.~Defrise, and C.~De~Mol, ``An iterative thresholding algorithm
  for linear inverse problems with a sparsity constraint,''
  \emph{Communications on Pure and Applied Mathematics}, vol.~57, no.~11, pp.
  1413--1457, 2004.

\bibitem{fista}
A.~Beck and M.~Teboulle, ``A fast iterative shrinkage-thresholding algorithm
  for linear inverse problems,'' \emph{SIAM J. Img. Sci.}, vol.~2, pp.
  183--202, March 2009.

\bibitem{Osher}
Y.~Li and S.~Osher, ``Coordinate descent optimization for $\ell_1$ minimization
  with application to compressed sensing; a greedy algorithm,'' \emph{Inverse
  Problems and Imaging}, vol.~3, pp. 487--503, 2009.

\bibitem{nesterov07}
Y.~Nesterov, ``Gradient methods for minimizing composite objective function,''
  in \emph{{CORE}}.\hskip 1em plus 0.5em minus 0.4em\relax Catholic University
  of Louvain, Louvain-la-Neuve, Belgium, 2007.

\bibitem{bertsekas1999nonlinear}
D.~Bertsekas, ``Nonlinear programming,'' 1999.

\bibitem{Bach11}
F.~Bach, R.~Jenatton, J.~Mairal, and G.~Obozinski, ``Convex optimization with
  sparsity-inducing norms,'' in \emph{Optimization for Machine Learning}.\hskip
  1em plus 0.5em minus 0.4em\relax MIT Press, 2011.

\bibitem{CaiCS10}
J.-F. Cai, E.~J. Cand{\`e}s, and Z.~Shen, ``A singular value thresholding
  algorithm for matrix completion,'' \emph{SIAM J. on Opt.}, vol.~20, no.~4,
  pp. 1956--1982, 2010.

\bibitem{MuDYY11}
Y.~Mu, J.~Dong, X.~Yuan, and S.~Yan, ``Accelerated low-rank visual recovery by
  random projection,'' in \emph{CVPR}, 2011, pp. 2609--2616.

\bibitem{Recht:2011wv}
B.~Recht and C.~R{\'e}, ``Parallel stochastic gradient algorithms for
  large-scale matrix completion,'' \emph{Optimization Online}, 2011.

\bibitem{Lin07projectedgradient}
C.-J. Lin, ``Projected gradient methods for non-negative matrix
  factorization,'' \emph{Neural Computation}, vol.~19, pp. 2756--2779, 2007.

\bibitem{colson2007overview}
B.~Colson, P.~Marcotte, and G.~Savard, ``An overview of bilevel optimization,''
  \emph{Annals of Operations Research}, vol. 153, no.~1, pp. 235--256, 2007.

\bibitem{jarrett2009best}
K.~Jarrett, K.~Kavukcuoglu, M.~Ranzato, and Y.~LeCun, ``What is the best
  multi-stage architecture for object recognition?'' in \emph{CVPR}, 2009, pp.
  2146--2153.

\bibitem{kavukcuoglu2010fast}
K.~Kavukcuoglu, M.~Ranzato, and Y.~LeCun, ``Fast inference in sparse coding
  algorithms with applications to object recognition,'' \emph{arXiv:1010.3467},
  2010.

\bibitem{LecunNN}
K.~Gregor and Y.~LeCun, ``Learning fast approximations of sparse coding,'' in
  \emph{ICML}, 2010, pp. 399--406.

\bibitem{hinton2006}
G.~Hinton and R.~Salakhutdinov, ``Reducing the dimensionality of data with
  neural networks,'' \emph{Science}, vol. 313, no. 5786, pp. 504--507, 2006.

\bibitem{autoencoders}
I.~Goodfellow, Q.~Le, A.~Saxe, H.~Lee, and A.~Y. Ng, ``Measuring invariances in
  deep networks,'' in \emph{In NIPS}, 2009, pp. 646--654.

\bibitem{Peng2011-PAMI}
Y.~Peng, A.~Ganesh, J.~Wright, W.~Xu, and Y.~Ma, ``{RASL}: Robust alignment by
  sparse and low-rank decomposition for linearly correlated images,'' in
  \emph{CVPR}, 2010, pp. 763--770.

\bibitem{ICML}
P.~Sprechmann, A.~M. Bronstein, and G.~Sapiro, ``Learning efficient structured
  sparse models,'' in \emph{ICML}, 2012.

\bibitem{SBS_ismir}
------, ``Real-time online singing voice separation from monaural recordings
  using robust low-rank modeling,'' in \emph{ISMIR}, 2012.

\bibitem{mairal2009online}
J.~Mairal, F.~Bach, J.~Ponce, and G.~Sapiro, ``Online dictionary learning for
  sparse coding,'' in \emph{ICML}, 2009, pp. 689--696.

\bibitem{ER10}
Y.~C. Eldar and H.~Rauhut, ``Average case analysis of multichannel sparse
  recovery using convex relaxation,'' \emph{{IEEE} Trans. on Inf. Theory},
  vol.~56, no.~1, pp. 505--519, 2010.

\bibitem{pca}
I.~T. Jolliffe, \emph{{Principal Component Analysis}}, 2nd~ed.\hskip 1em plus
  0.5em minus 0.4em\relax Springer, 2002.

\bibitem{mateos-2011}
G.~Mateos and G.~B. Giannakis, ``Robust {PCA} as bilinear decomposition with
  outlier-sparsity regularization,'' \emph{arXiv.org:1111.1788}, 2011.

\bibitem{morteza}
M.~Mardani, G.~Mateos, and G.~B. Giannakis, ``Unveiling network anomalies in
  large-scale networks via sparsity and low rank,'' in \emph{Proc. of 44th
  Asilomar Conf. on Signals, Systems, and Computers}, 2011.

\bibitem{friedman10a}
J.~Friedman, T.~Hastie, and R.~Tibshirani, ``A note on the group lasso and a
  sparse group lasso,'' 2010, preprint.

\bibitem{Tseng}
P.~Tseng, ``Convergence of a block coordinate descent method for
  nondifferentiable minimization,'' \emph{J. Optim. Theory Appl.}, vol. 109,
  no.~3, pp. 475--494, June 2001.

\bibitem{Vapnik71}
V.~Vapnik and A.~Chervonenkis, ``On the uniform convergence of relative
  frequencies of events to their probabilities,'' \emph{Theory of Probability
  \& Its Applications}, vol.~16, no.~2, pp. 264--280, 1971.

\bibitem{bottou-2010}
L.~Bottou, ``{Large-scale machine learning with stochastic gradient descent},''
  in \emph{COMPSTAT}, August 2010, pp. 177--187.

\bibitem{denoising-2012}
P.~Sprechmann, A.~M. Bronstein, M.~M. Bronstein, and G.~Sapiro, ``Learnable low
  rank sparse models for speech denoising,'' \emph{arXiv.org:1221.1288}, 2012.

\bibitem{brodatz}
T.~Randen and J.~H. Husoy, ``Filtering for texture classification: a
  comparative study,'' \emph{IEEE Trans. Pattern Anal. Mach. Intell.}, vol.~21,
  no.~4, pp. 291--310, 1999.

\bibitem{audio}
P.~Sprechmann, I.~Ramirez, P.~Cancela, and G.~Sapiro, ``Collaborative sources
  identification in mixed signals via hierarchical sparse modeling,'' in
  \emph{Proc. ICASSP}, May 2011.

\bibitem{hsu2010improvement}
C.~Hsu and J.~Jang, ``On the improvement of singing voice separation for
  monaural recordings using the {MIR-1K} dataset,'' \emph{IEEE Trans. on Audio,
  Speech, and Lang. Proc.}, vol.~18, no.~2, pp. 310--319, 2010.

\bibitem{vincent2006performance}
E.~Vincent, R.~Gribonval, and C.~F{\'e}votte, ``Performance measurement in
  blind audio source separation,'' \emph{IEEE Trans. on Audio, Speech, and
  Lang. Proc.}, vol.~14, no.~4, pp. 1462--1469, 2006.

\bibitem{cooke2006audio}
M.~Cooke, J.~Barker, S.~Cunningham, and X.~Shao, ``An audio-visual corpus for
  speech perception and automatic speech recognition,'' \emph{J. of the Acoust.
  Society of America}, vol. 120, p. 2421, 2006.

\bibitem{PearceH00}
D.~Pearce and H.-G. Hirsch, ``The {AURORA} experimental framework for the
  performance evaluation of speech recognition systems under noisy
  conditions,'' in \emph{INTERSPEECH}, 2000, pp. 29--32.

\bibitem{nam2012cosparse}
S.~Nam, M.~Davies, M.~Elad, and R.~Gribonval, ``The cosparse analysis model and
  algorithms,'' \emph{Applied and Computational Harmonic Analysis}, 2012.

\bibitem{vaiter2011robust}
S.~Vaiter, G.~Peyr{\'e}, C.~Dossal, and J.~Fadili, ``Robust sparse analysis
  regularization,'' \emph{arXiv preprint arXiv:1109.6222}, 2011.

\end{thebibliography}

\end{document}